\documentclass{article}

\usepackage[utf8]{inputenc} % allow utf-8 input
\usepackage[T1]{fontenc}    % use 8-bit T1 fonts
\usepackage{url}            % simple URL typesetting
\usepackage{booktabs}       % professional-quality tables
\usepackage{multicol}
\usepackage{multirow}
\usepackage[margin=1.0in]{geometry}
\usepackage{amsfonts}       % blackboard math symbols
\usepackage{nicefrac}       % compact symbols for 1/2, etc.
\usepackage{microtype}      % microtypography
\usepackage{xcolor}
\usepackage{amsmath}
\usepackage{amsthm}
\usepackage{mathtools}
\usepackage{lineno}
\usepackage{marvosym}

\usepackage{array}
\newcolumntype{L}[1]{>{\raggedright\arraybackslash}p{#1}}
\newcolumntype{C}[1]{>{\centering\arraybackslash}p{#1}}

\theoremstyle{definition}

\usepackage{xr-hyper}
\makeatletter
\newcommand*{\addFileDependency}[1]{% argument=file name and extension
  \typeout{(#1)}
  \@addtofilelist{#1}
  \IfFileExists{#1}{}{\typeout{No file #1.}}
}
\makeatother

\usepackage{graphicx}
\usepackage[style=nature]{biblatex}
\usepackage{algorithm}
\usepackage[noend]{algpseudocode}
\usepackage{amsmath}
\usepackage{amsthm}
\usepackage{enumitem}
\usepackage{authblk}
\algnewcommand{\IIf}[1]{\State\algorithmicif\ #1\ \algorithmicthen}
\algnewcommand{\EndIIf}{\unskip\ \algorithmicend\ \algorithmicif}

\DeclareMathOperator*{\argmin}{arg\,min}

\usepackage{todonotes}
\usepackage{datenumber}
\usepackage{eqname}
\usepackage{float}
\usepackage{setspace}
\usepackage{longtable}
\usepackage{hyperref}

\DeclareFixedFont{\ttb}{T1}{txtt}{bx}{n}{9} % for bold
\DeclareFixedFont{\ttm}{T1}{txtt}{m}{n}{9}  % for normal
\usepackage{color}
\definecolor{forestgreen}{rgb}{.133, .545, .133}
\definecolor{deepblue}{rgb}{0.255,0.412,0.882}
\definecolor{deepred}{rgb}{0.6,0,0}
\definecolor{deepgreen}{rgb}{0,0.5,0}

\usepackage{tcolorbox}

\title{Algorithms to estimate Shapley value feature attributions}
\author[1,*]{Hugh Chen}
\author[1,*]{Ian C. Covert}
\author[2]{Scott M. Lundberg}
\author[1,$\dagger$]{Su-In Lee}

\affil[1]{{\small Paul G. Allen School of Computer Science and Engineering, University of Washington}}
\affil[2]{{\small Microsoft Research}}
\affil[*]{{\small Equal contribution}}
\affil[$\dagger$]{{\small Corresponding: suinlee@cs.washington.edu}}

\begin{document}

\setcounter{page}{1}

\date{}

{\setstretch{1}
\maketitle
}

\begin{abstract}
Feature attributions based on the Shapley value are popular for explaining
% a
% very
% popular way to explain
machine learning models; however, their estimation is complex from both a theoretical and computational standpoint.
% both theoretically and computationally.  
% this complexity has led to the development of a wide diversity of algorithms.  
We disentangle this complexity into two factors: (1)~the approach to removing feature information,
% remove features
and (2)~the tractable estimation strategy.  These two factors provide a natural lens through which we can better understand
% comprehend
and compare 24 distinct algorithms.
% Based on these factors, we clarify important, but often underappreciated parameters of these techniques.  
% To our knowledge, we provide the first survey of algorithms to estimate Shapley value feature attributions. 
Based on the various feature removal approaches, we describe the multiple types of Shapley value feature attributions and methods to calculate
% estimate
each one. Then, based on the tractable estimation strategies, we characterize two distinct families of approaches: model-agnostic and model-specific approximations.  For the model-agnostic approximations,
% estimators,
we benchmark a wide class of estimation
% approximation
approaches and tie them to alternative yet equivalent characterizations of the Shapley value.  For the model-specific approximations,
% estimators,
we clarify the assumptions crucial to each method's tractability for linear, tree, and deep models.  Finally, we identify gaps in the literature and promising future research directions.
\end{abstract}

\section{Introduction}
\label{sec:introduction}

% \begin{itemize}
%     \item Include visualization of comparison of approximation approaches
%     \item Think about how to improve discussion of the assumption based approaches
% \end{itemize}

Machine learning models are increasingly prevalent because they have matched or surpassed human performance
% humans
in many applications: these include Go~\cite{silver2017mastering}, poker~\cite{moravvcik2017deepstack}, Starcraft~\cite{vinyals2019grandmaster}, protein folding~\cite{jumper2021highly}, language translation~\cite{jean2014using}, and more.  One critical component in their success is flexibility, or expressive power~\cite{breiman2001random,lecun2015deep,chen2016xgboost}, which has been facilitated by more complex models and improved hardware
% recent technological developments
\cite{steinkraus2005using}.  Unfortunately, their flexibility also makes models opaque, or challenging for humans to understand.
% hard to understand from the perspective of humans.
Combined with the tendency of machine learning to rely on shortcuts~\cite{geirhos2020shortcut} (i.e., unintended learning strategies that fail to generalize to unseen data), there is a growing demand for model interpretability~\cite{doshi2017towards}.  This demand is reflected in increasing
% an increase in
calls for explanations by diverse regulatory bodies, such as the General Data Protection Regulation's ``right to explanation''~\cite{selbst2018meaningful} and the Equal Credit Opportunity Act's adverse action notices~\cite{knight2019ai}.

There are many possible ways to explain machine learning models (e.g., counterfactuals, exemplars, surrogate models, etc.), but one extremely popular approach
% class of explanation
is \textit{local feature attribution}. In this approach, individual predictions are explained by
% explanations of a model's prediction for a specific sample are given by
% an importance vector
an attribution vector $\phi \in \mathbb{R}^d$, with $d$ being the number of features used by the model.  One prominent example is LIME~\cite{ribeiro2016should}, which fits
% explains a machine learning model by fitting
% training
a simple interpretable model that captures
% aims to capture
the model's behavior in the neighborhood of a single sample; when a linear model is used, the coefficients serve as attribution scores for each feature.
% sample's prediction.
% The most common choice is a linear model whose
% % where the
% coefficients determine the explanation, thereby providing attribution values for each feature.
In addition to LIME, many other methods exist to compute local feature attributions~\cite{ribeiro2016should,lundberg2017unified,lundberg2020local,shrikumar2017learning,binder2016layer,datta2016algorithmic,sundararajan2017axiomatic}.  One popular class of approaches
% category of approaches
is \textit{additive feature attribution methods}, which are those whose attributions sum to
a specific value, such as
% : these are local feature attributions that sum to a specific value, often
% such as
the model's prediction~\cite{lundberg2017unified}.
% (analogous to the efficiency axiom in Figure~\ref{fig:shapley_values}b).

To unify the class of additive feature attribution methods, \textcite{lundberg2017unified} introduced SHAP
% ~\cite{lundberg2017unified}
as a unique solution determined by additional desirable properties (Section~\ref{sec:shapley_values}).
% (Figure~\ref{fig:shapley_values}).
% However, this
Its uniqueness depends on defining a coalitional game (or set function) based on the model being explained (a connection first introduced in \cite{strumbelj2010efficient}).
% but models are not naturally equivalent to coalitional games.
% machine learning models.
% In particular,
\textcite{lundberg2017unified}
% \citeauthor{lundberg2017unified}
initially defined the game as the expectation of the model's output
% a conditional expectation of the model's output,
when conditioned on a set of observed features.
However, given the difficulty of computing conditional expectations in practice, the authors suggested using a marginal expectation that ignores dependencies between the observed and unobserved features.
% where dependencies between the observed features and unobserved features are ignored.
This point of complexity has led to distinct Shapley value approaches that differ in 
% the development of distinct types of feature attributions based on
how they remove features~\cite{kumar2020problems,sundararajan2020many,janzing2020feature,heskes2020causal,covert2021explaining}, as well as subsequent interpretations of how these
% different
two approaches relate to causal interventions~\cite{janzing2020feature, heskes2020causal} or information theory~\cite{chen2018shapley, covert2021explaining}.  Moving forward, we will refer to all feature attributions based on the Shapley value as \textit{Shapley value explanations}.

Alongside the definition of the coalitional game, another
% the other
% difficulty
challenge for Shapley value explanations is that 
calculating them
% exactly
% their exact calculation
has computational complexity that is exponential in the number of features.
% exponential computational complexity.
%~\cite{van2021tractability}.
The original SHAP paper~\cite{lundberg2017unified} therefore discussed several strategies for approximating Shapley values, including weighted linear regression (KernelSHAP~\cite{lundberg2017unified}), sampling feature 
% permutations
combinations
(IME~\cite{strumbelj2010efficient}), and several model-specific approximations (LinearSHAP~\cite{lundberg2017unified,chen2020true}, MaxSHAP~\cite{lundberg2017unified}, DeepSHAP~\cite{lundberg2017unified,chen2021explaining}).  Since the original work, other methods have been developed to estimate Shapley value explanations more efficiently, using model-agnostic strategies (permutation~\cite{castro2009polynomial}, multilinear extension~\cite{okhrati2021multilinear}, FastSHAP~\cite{jethani2021fastshap})
% , as well as
and
model-specific strategies (linear models~\cite{chen2020true}, tree models~\cite{lundberg2020local}, deep models~\cite{chen2021explaining,ancona2019explaining,wang2020shapley}).  Of these two categories, model-agnostic approaches are more flexible but stochastic, whereas model-specific approaches are
% less flexible but
significantly faster to calculate.
% In order to
To better understand the model-agnostic approaches, we
% categorize the approximation algorithms based on equivalent mathematical definitions of the Shapley value
% describe
present a categorization of the approximation algorithms
% approaches
based on equivalent mathematical definitions of the Shapley value, and we empirically compare their convergence properties
(Section~\ref{sec:shapley_value_feature_attributions}).  Then, to better understand the model-specific approaches, we highlight the key assumptions underlying each approach (Section~\ref{sec:shapley_explanation_algorithms}).

\begin{figure}[!ht]
\includegraphics[width=\textwidth]{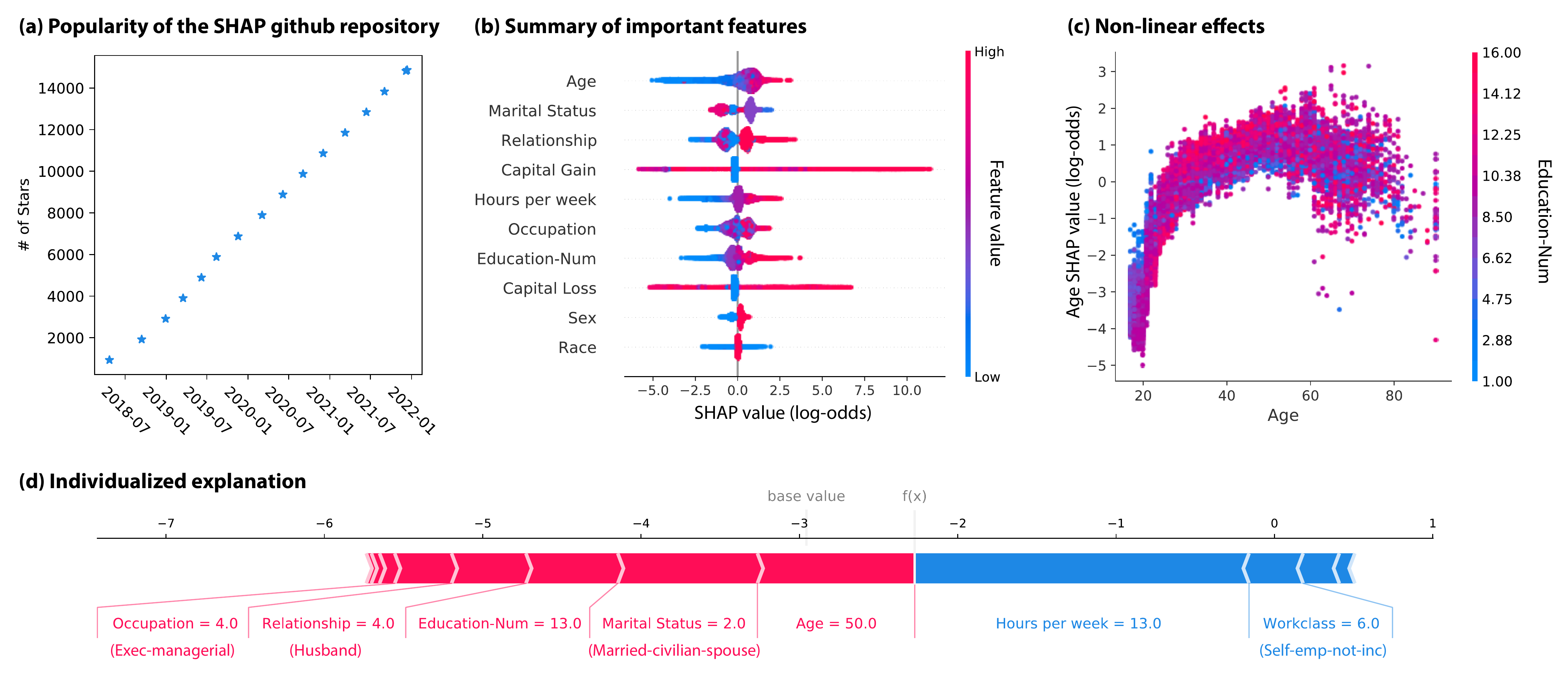}
\centering
\caption{Shapley value explanations
% feature attributions
are popular and practical.  (a)~The large number of Github stars on shap (\url{https://github.com/slundberg/shap}), the most famous package to estimate Shapley value explanations,
indicates their popularity.  (b)-(d)~A real-world example of Shapley value explanations
% feature attributions
for a tree ensemble model trained to predict whether individuals have income greater than 50,000 dollars based on census data.  (b)~Local feature attributions enable a global understanding of important features.  (c)~Local feature attributions help explain non-linear and interaction effects.  (d)~Local feature attributions explain how an individual's features influence their outcome.}
\label{fig:concept}
\end{figure}

These two sources of complexity, properly removing features and accurately approximating Shapley values, have led to a wide variety of papers and algorithms on the subject.  Unfortunately, this abundance of algorithms coupled with the inherent complexity of the topic have made the literature difficult to navigate, which can lead to misuse, especially given the popularity of Shapley value explanations (Figure~\ref{fig:concept}a).  To address this, we provide an approachable explanation of the sources of complexity underlying the computation of Shapley value explanations.  

We discuss these difficulties in detail, beginning by introducing the preliminary concepts of feature attribution (Section~\ref{sec:feature_attributions}) and the Shapley value (Section~\ref{sec:shapley_values}).  Based on the various feature removal approaches, we then describe popular variants of Shapley value explanations
% in addition to
as well as
approaches to estimate the corresponding coalitional games (Section~\ref{sec:shapley_value_feature_attributions}).
% each one's coalitional game. 
Next, based on the estimation strategies, we describe model-agnostic and model-specific algorithms that rely on approximations and/or assumptions to tractably estimate Shapley value explanations (Section \ref{sec:shapley_explanation_algorithms}).  These two sources of complexity provide a natural lens through which we present what is, to our knowledge, the first
% the first, to our knowledge,
comprehensive survey of 24 distinct algorithms\footnote{This count excludes minor variations of these algorithms.} that combine different feature removal and tractable estimation strategies to compute Shapley value explanations.
% , combining different feature removal and tractable estimation strategies.
  Finally, we identify gaps and important future directions in this area of research throughout the article.
% research directions in this domain.

\section{Feature attributions}
\label{sec:feature_attributions}

Given a model $f$ and features $x_1, \ldots, x_d$, feature attributions explain predictions by assigning
% Feature attributions explain the prediction for a model $f$ by assigning
scalar values that represent each feature's importance.  For an intuitive description of feature attributions, we first consider linear models.  Linear models of the form $f(x)=\beta_0 +\beta_1x_1 +\cdots + \beta_d x_d$ are often considered interpretable because each feature is linearly related to the prediction
% model output
via a single parameter.  In this case, a common \textit{global feature attribution} that describes the model's overall dependence on feature $i$ is the corresponding
% model
coefficient $\beta_i$. 
% $\phi$ that explains the model as a whole is the slope $\beta_i$. 
For linear models, each coefficient $\beta_i$ describes the influence that variations in feature $x_i$ have on the  model output.
% For linear models, the slope is an intuitive global attribution because it describes the relationship between feature $i$ and the model's output for any value of $x_i$.

Alternatively, it may be preferable to give an individualized explanation that is not for the model as a whole, but rather for the prediction
% output
$f(x^e)$ given a specific sample $x^e$.  These types of explanations are known as \textit{local feature attributions}, and the sample being explained ($x^e$) is called the \textit{explicand}. For linear models, one reasonable local feature attribution is $\phi_i(f,x^e)=\beta_ix^e_i$, because it is exactly the contribution that feature $i$ makes to the model's prediction
% output
for the given explicand.  However, note that this attribution hides within it an implicit assumption that we want to compare against an alternative feature value
% may not capture the whole story; it implicitly considers an alternative
of $x_i = 0$, but we may wish to account for other plausible alternative values, or more generally for the feature's distribution
% \footnote{Note that the local feature attribution of $\beta_ix^e_i$ is actually the marginal Shapley value for linear models if each feature has already been standardized to zero mean and unit variance (which is a common practice for lasso or ridge regression).}
or statistical relationships with other features (Section~\ref{sec:shapley_value_feature_attributions}).
% However, as a whole, linear models are generally considered interpretable because they can be understood, to a certain extent, through each feature's coefficients.

Linear models offer a simple case where we can understand each feature's role via the model parameters,
% model's coefficients,
but this approach does not extend naturally to more complex model types. For model types that are most widely used today, including tree ensembles and deep learning models, their large number of operations prevents us from understanding each feature's role by examining the model parameters. These flexible, non-linear models can capture more patterns in data, but they require us to develop more sophisticated and generalizable notions of feature importance. 
% Recently,
Thus, many researchers
% now rely on
% have thus turned to
have
recently
begun turning to
% Although interpretable models are appealing, they often require unrealistic assumptions (e.g., linear relationships, limited interaction effects). Instead, flexible non-linear models (e.g., tree or deep models) can capture more complex patterns in data. Unfortunately, complex models cannot be examined as easily, because each parameter in a complex model is no longer associated with a single feature.  In order to assign importance in a more principled manner, many researchers have
Shapley value explanations
% feature attributions,
% which can
to summarize important features (Figure~\ref{fig:concept}b), surface non-linear effects (Figure~\ref{fig:concept}c), and provide individualized explanations (Figure~\ref{fig:concept}d) in an axiomatic manner (Figure~\ref{fig:shapley_values}b).

% For these models, summarizing non-linear effects with a single number (a global feature attribution $\phi_i(f)$) is much harder, because the slope for any feature changes based on that feature's value and also based on other features' values. Instead, we can aim for local feature attributions ($\phi_i(f,x^e)$) that will explain the model’s prediction for a specific individual. In order to do so in a principled manner, we can turn to the Shapley values.

\section{Shapley values}
\label{sec:shapley_values}

\begin{figure}[!ht]
\includegraphics[width=\textwidth]{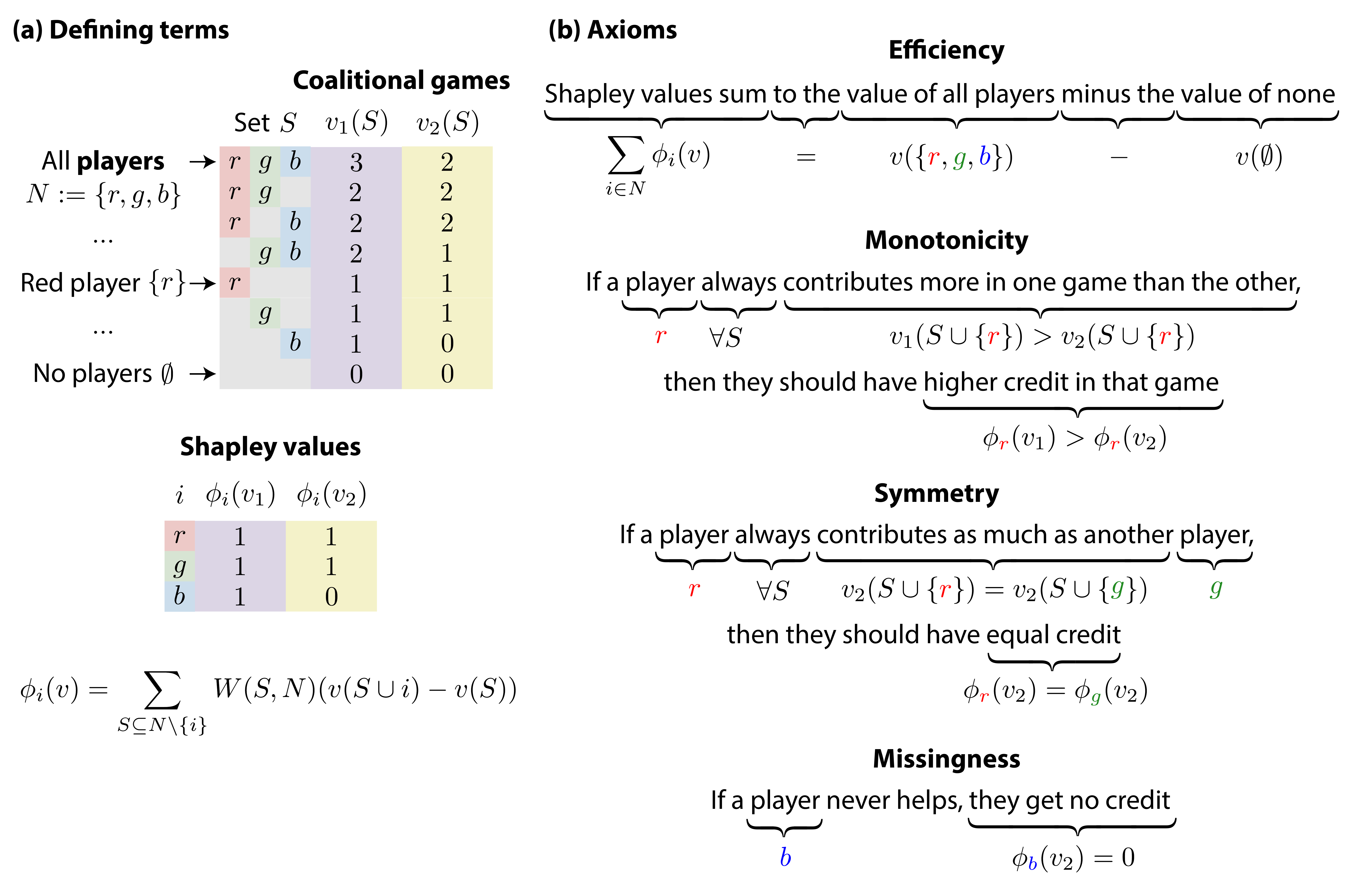}
\centering
\caption{(a)~Defining terms related to the Shapley value.  Players either participate or abstain from the coalitional game, and the game  
% Coalitional games map
maps from any subset of participating players to a scalar value.  Shapley values are a solution concept to allocate
% spread
credit to each player in
% based on
a coalitional game.  (b)~A sufficient, but not exhaustive set of axioms that uniquely define the Shapley value.}
\label{fig:shapley_values}
\end{figure}

Shapley values are a tool from game theory~\cite{shapley1953value} designed to allocate
% spread
credit to
% among
players in coalitional games.  The \textit{players} are represented by a set $D=\{1,\ldots,d\}$, and the \textit{coalitional game} is a function that maps from
% all
subsets of the players to a scalar value. A game is represented by a subset function $v(S):\mathcal{P}(D)\mapsto\mathbb{R}$, where $\mathcal{P}(D)$ is the power set of $D$ (representing all possible subsets of players) (Figure~\ref{fig:shapley_values}a).

To make these concepts more concrete, we can imagine a company that makes a profit $v(S)$
% that is
determined by the set of employees $S \subseteq D$ that choose to work that day. A natural question is how to compensate the employees for their contribution to the total profit. Assuming we know the profit for all subsets of employees, Shapley values assign credit to an individual $i$ by calculating
% taking
a weighted average of the profit increase when $i$ works with group $S$ versus when $i$ does not work with group $S$ (the marginal contribution). Averaging this difference over all possible subsets $S$ to which $i$ does not belong ($S \subseteq D \setminus \{i\}$), we arrive at the definition of the Shapley
value:
% (Figure~\ref{fig:shapley_values}a):
\begin{align}
\overbrace{\phi_i(v)}^{i\text{'s Shapley value}}=\sum_{S\subseteq D\setminus\{i\}}\underbrace{\frac{|S|!(|D|-|S|-1)!}{|D|!}}_{S\text{'s weight}}(\overbrace{v(S\cup\{i\})-v(S)}^{i\text{'s marginal contribution}}) \label{eq:shapley_definition}
\end{align}

Shapley values offer a compelling way to spread credit in coalitional games, and they have been widely adopted in fields including computational biology~\cite{lucchetti2010shapley,moretti2010statistical}, finance~\cite{tarashev2016risk,tarashev2009systemic}, and more~\cite{landinez2017shapley,aumann1994economic}.  Furthermore, they are a unique solution to the credit allocation
% spreading
problem as defined by several desirable properties~\cite{shapley1953value,young1985monotonic} (Figure~\ref{fig:shapley_values}b).

% \section{Shapley value feature attributions}
\section{Shapley value explanations}
\label{sec:shapley_value_feature_attributions}

In this section, we present common strategies to define local feature attributions based on the Shapley value. We also present intuitive examples based on explaining linear models, and we discuss the
% a few
tradeoffs between various approaches to removing features.

% \textcolor{blue}{Probably good to have a sentence here introducing or summarizing the section.}

\subsection{Machine learning models are not coalitional games}

\begin{figure}[!ht]
\includegraphics[width=\textwidth]{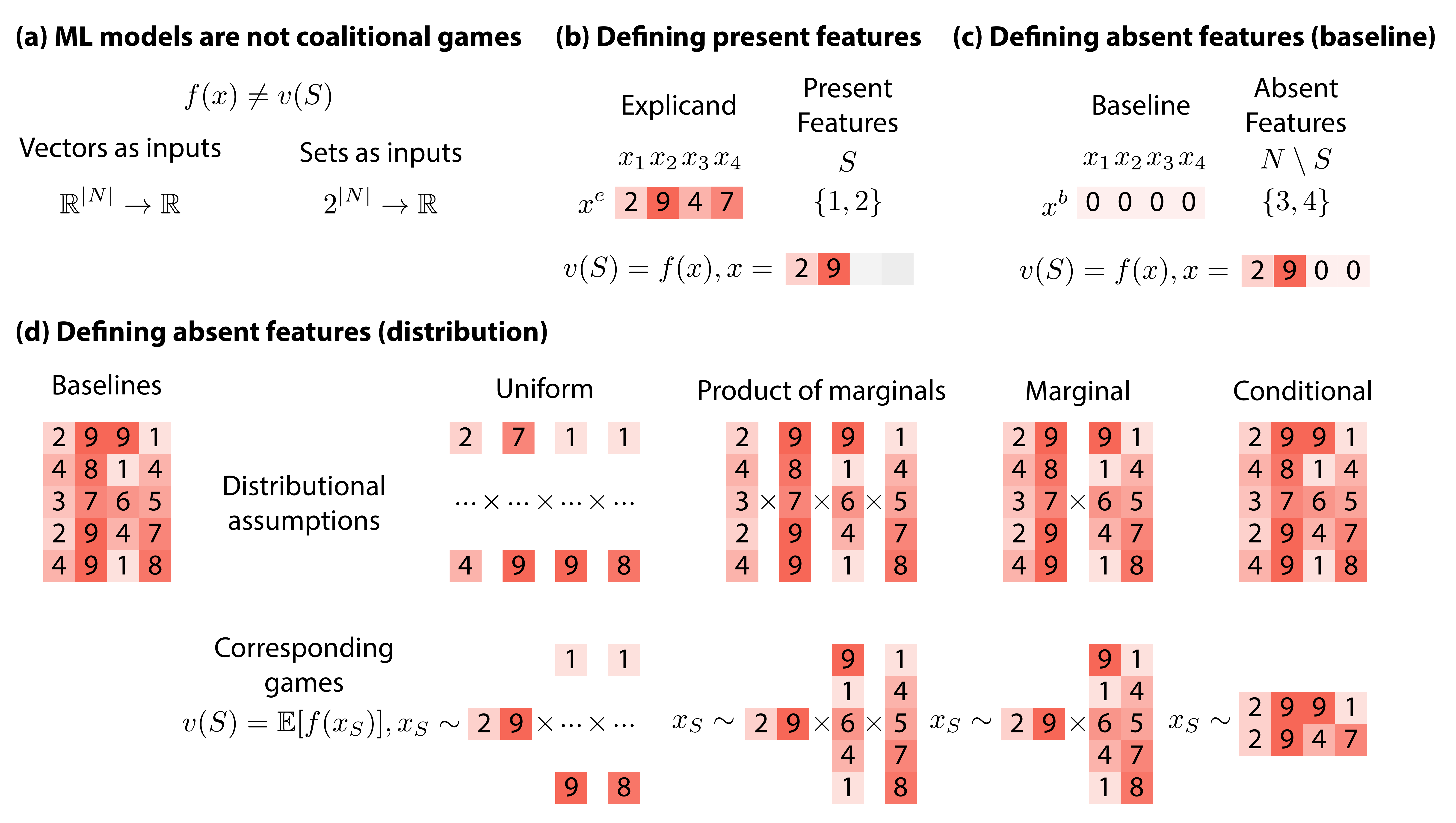}
\centering
\caption{Empirical strategies for handling absent features.  (a) Machine learning models have vector inputs and coalitional games have set inputs.  For simplicity of notation we assume real-valued features, but Shapley value explanations can accommodate discrete features (unlike gradient-based methods).  (b) Present features are replaced according to the explicand.  (c) Absent features can be replaced according to a baseline.  (d) Alternatively, absent features can be replaced according to a set of baselines with different distributional assumptions.  In particular, the uniform approach uses the range of the baselines' absent features to define independent uniform distributions to draw absent features from.  The product of marginals approach draws each absent feature independently according to the values seen in the baselines.  The marginal approach draws groups of absent feature values that appeared in the baselines.  Finally, the conditional approach only considers samples that exactly match on the present features.  Note that this figure depicts empirically estimating each expectation; however, in practice, the conditional approach is estimated by fitting models (Section~\ref{sec:conditional_shapley}).}
\label{fig:absent_features}
\end{figure}

Although Shapley values are an attractive
% optimal
solution for allocating credit among players in coalitional games, our goal is to allocate credit among features $x_1,\ldots,x_d$
% ($x:=\{x_1,\ldots,x_d\}\in\mathbb{R}^d$)
in a machine learning model $f(x) \in \mathbb{R}$.
% ($f(x): \mathbb{R}^d \mapsto \mathbb{R}$),
% which is generally not a coalitional game.
% Importantly,
Machine learning models are not coalitional games by default, so to use Shapley values
% to explain a model,
\textbf{we must first define a coalitional game $v(S)$ based on the model $f(x)$} (Figure~\ref{fig:absent_features}a).
The coalitional game can be chosen to represent various model behaviors, including the model's loss for a single sample or for the entire dataset~\cite{covert2021explaining}, but our focus is the most common choice:
% --
explaining the prediction $f(x^e)$ for a single sample $x^e$. 
% In order to define this coalitional game, we often need to be clear about what aspect of model behavior we are explaining.  Some popular choices of model behavior to explain include a single sample's prediction (local), a single sample's loss (local), or the model's loss on an entire data set (global)~\cite{covert2021explaining}.  In this paper, we will address local feature attributions, which have been the focus of many recent papers.

% As a first step, we can look at a machine learning model that \textit{actually is a coalitional game}: a machine learning model with binary features $x_i\in(0,1)$\footnote{Where 1 indicates presence of a feature and 0 indicates absence of a feature.}.  For such a model, the most natural coalitional game is the function applied to a vector that is one if the feature is in the set $S$ and zero otherwise: $v(S):=f(z^S)\text{, where } z^S_i = 1\text{ if }i\in S\text{, }0\text{ otherwise}$.  We can easily use Shapley values to explain this game and obtain feature attributions.

When explaining
% In order to explain
a machine learning model, it is natural to view each feature $x_i$ as a player in the coalitional game. However, we then must define what is meant by the presence or absence of each feature.
% features\footnote{More accurately, we have to define a new coalitional game $v(S)$ related to the model's prediction $f(x)$ where the features in $S$ are present and the remaining features are absent.}.
Given our focus on a single explicand $x^e$, the presence of feature $i$ will mean that the model is evaluated with the observed value $x^e_i$
(Figure~\ref{fig:absent_features}b).
As for the absent features, we next consider how to remove them to properly
% Next, we consider how to handle the absent features, which we must
% % want to
% remove 
% % that differ from those in the explicand
% to
assess the influence of the present features.
% First, to address the presence of a feature, our goal is to obtain \textit{local feature attributions}: a vector of importances for a specific sample $x^e$'s prediction (explicand).  So if feature $i$ is present, we can simply set that feature to be from the explicand ($x_i^e$) (Figure~\ref{fig:absent_features}b). The next step is to address the absence of a feature which has been approached in many ways.

% \subsection{Handling absent features (baseline)}
\subsection{Removing features with baseline values}
\label{sec:masking_missing}

\begin{figure}[!ht]
\includegraphics[width=\textwidth]{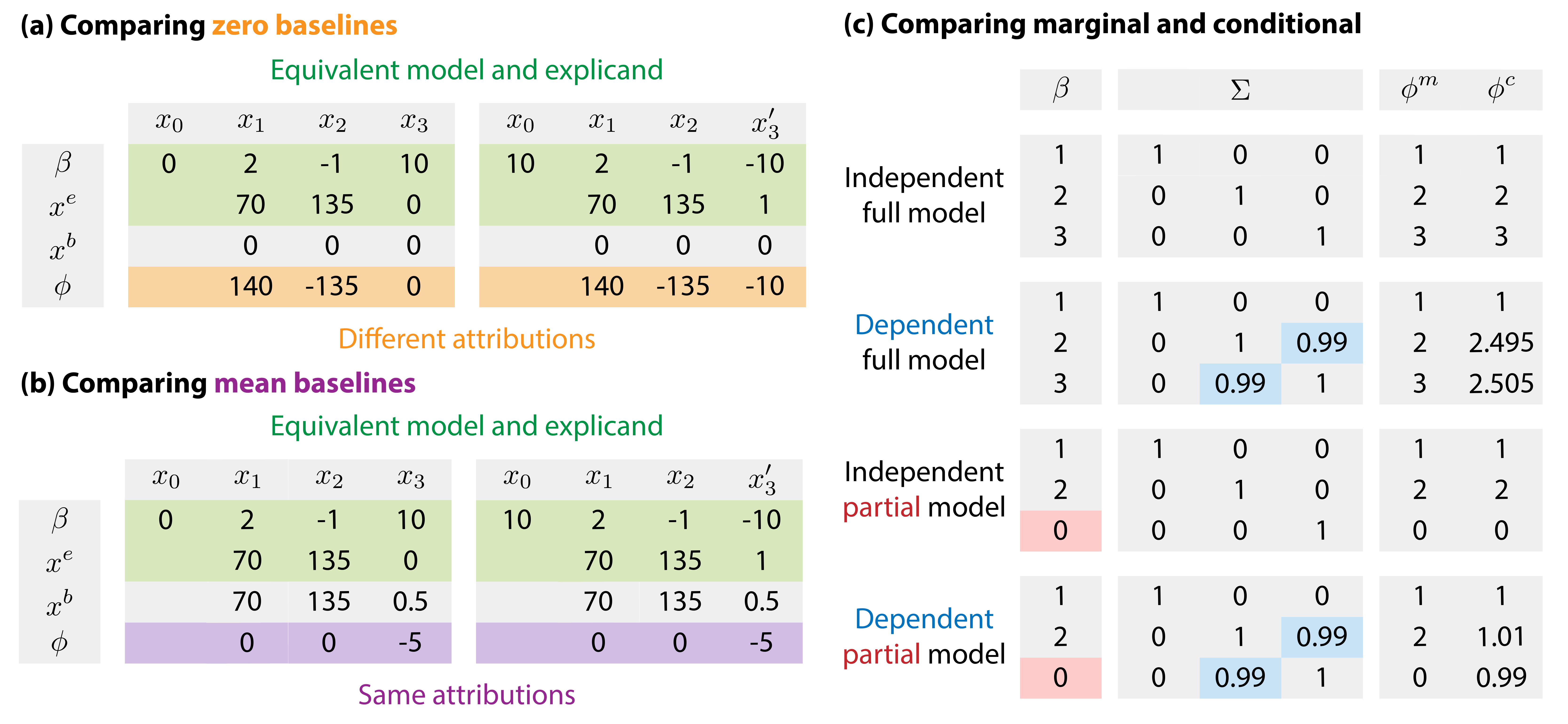}
\centering
\caption{Shapley values for linear models.  (a)-(b) A linear model ($\beta$), an explicand ($x^e$), a baseline ($x^b$), and baseline Shapley values ($\phi$) where feature 1 represents height (inches), feature 2 represents weight (lbs), and feature 3 represents gender.  Features $x_3$ and $x_3'$ denote different ways to represent gender, where $x_3=1$ is male and $x_3'=1$ is female.  (a) The models and explicands on the left and right are equivalent, but a zero baseline has a different meaning in each example and thus produces different attributions. (b)  In this case, we use an mean baseline, for which the encoding of gender does not affect the baseline Shapley values.  (c) Comparing marginal and conditional Shapley values for different models and feature dependencies with explicand $x^e=(1,1,1)$ and baseline $x^b=(0,0,0)$.  Vectors $\beta$ (linear model coefficients), $\phi^m$ (marginal Shapley values), and $\phi^c$ (conditional Shapley values) have elements corresponding to $x_1, x_2, x_3$, and matrix $\Sigma$'s columns and rows are $x_1, x_2, x_3$.  The independent models have no correlation between features and the dependent models have a surrogate feature (a highly correlated pair of features).  The full model has all non-zero coefficients whereas the partial model has a zero coefficient for the third feature.}
\label{fig:linear_example}
\end{figure}

% \caption{Example illustrating the downsides of a single baseline for Shapley values.  Feature 1 corresponds to age (inches), feature 2 is weight (pounds), and feature 3 is gender (0 is male, 1 is female).  The first baseline is an all-zero baseline.  The second is an average feature value baseline.  See Appendix Section~\ref{sec:masking_missing_proof} for the derivation of the Shapley values for a linear model for masking.}
% \label{tab:baseline}
% \end{figure}

% \caption{Comparing two versions of Shapley values: conditional expectation ($\phi^{CE}$) and marginal expectation ($\phi^{ICE}$). We make two simplifying assumptions: 1.) The function is linear ($y=\beta^T x$) and 2.) the data is multivariate normally distributed ($x\sim \mathcal{D}_3(0,\Sigma)$).  See Appendix Section~\ref{sec:ice_ce_proof} for the derivation of the Shapley values for ICE and CE.}
% \label{fig:interventional_vs_conditional}

One straightforward way to remove a feature is to replace its value using
% according to
a baseline sample $x^b$. That is, if a feature $i$ is absent, we simply set that feature's value to be $x_i^b$.  Then, the coalitional game is
% simply
defined as $v(S)=f(\tau(x^e, x^b, S))$, where we define $\tau(x^e, x^b, S)_i = x^e_i\text{ if }i\in S$ or $x^b_i$ otherwise (Figure~\ref{fig:absent_features}c).  In words, we evaluate the model on a new sample
where present features are the explicand values and absent features are the baseline values.
% by where absent features are the baseline values and present features are the explicand values.
As shorthand notation, we will
% use $f(x^e_S, x^b_{\bar{S}})$ to
refer to $f(\tau(x^e, x^b, S))$
as $f(x^e_S, x^b_{\bar{S}})$ 
in the remainder of the paper.
% future Sections.
% and present features using the explicand.

The Shapley values for this coalitional game are referred to as \textit{baseline Shapley values}
% , and has been shown to be a unique solution to attribution methods with a single baseline based on cost-sharing literature
\cite{sundararajan2020many}.  This approach is simple to implement, but
% However,
the choice of the baseline is not straightforward and can be somewhat arbitrary.
% As evidence,
Many different baselines have been considered, including an all-zeros baseline, an average across features\footnote{This baseline is most natural for image data~\cite{fong2017interpretable,sturmfels2020visualizing}.},
% \footnote{This baseline was used to remove high frequency signals using a Gaussian kernel for image data in particular~\cite{fong2017interpretable,sturmfels2020visualizing}.}
a baseline drawn from a uniform distribution, and more~\cite{sundararajan2017axiomatic,shrikumar2017learning,fong2017interpretable,sturmfels2020visualizing,kapishnikov2019xrai,ren2021learning}.  Unfortunately, the choice of baseline
% different baseline choices choices of baseline
heavily influences the feature attributions, and the criteria for choosing
% .  In general, the motivation to choose
a baseline can be unclear. One possible motivation could be to find a neutral, uninformative
% uninteresting
baseline, but such a baseline value may not exist.  For these reasons, it is common to use a distribution of baselines instead of
% heavily
relying on a single baseline.

\subsection{Removing features with distributional values}
\label{sec:impute_conditioning}

% One natural approach is to define the coalitional game as an average of the model's prediction based on values sampled from a distribution.

Rather than setting the removed features to fixed baseline values, another option is to average the model's prediction across randomly sampled replacement values; this may offer a better method to represent
% handle
absent feature information.
% values.
% Considering multiple values may offer a better approach to represent the absence of information.
% from the removed features.
A first approach is to sample from the \textit{conditional distribution} for the removed features. That is,
given an explicand $x^e$ and subset $S \subseteq D$,
we can consider the set of present features $x_S^e$
% assume we have access to the conditional distribution $p(x_{\bar S} \mid x_S)$ for the absent features, and then sample multiple replacement values according to $x_{\bar S} \sim p(x_{\bar S} \mid x_S^e)$.
and then sample replacement values for the absent features according to $x_{\bar S} \sim p(x_{\bar S} \mid x_S^e)$.
In this case, the coalitional game is defined as the expectation of the prediction $f(x^e_S, x_{\bar S})$ across this distribution.
% The first natural approach that removes features according to a baseline distribution uses a conditional distribution.  Instead of simply replacing absent features with a fixed value, we condition on the set of features that are present ($S$).  Then, conditioning on $S$ indicates that we know features in $S$ are from the explicand and we use those feature values to sample plausible (on-manifold) values for the absent features.  Finally, the coalitional game for subset $S$ is the conditional expectation: the average of the model's prediction based on these on-manifold samples.
% If we define $\mathcal{D}$ to be the baseline distribution our samples are drawn from, the value of the game can be based on a conditional expectation
% $v(S)=\mathbb{E}_\mathcal{D}[f(x)\mid x_{S}]$
% , where conditioning on $S$ indicates that we know features in $S$ are from the explicand.  
There are several names for Shapley values with this coalitional game: observational Shapley values~\cite{chen2020true}, conditional expectation Shapley~\cite{sundararajan2020many}, and finally \textit{conditional Shapley values}~\cite{heskes2020causal}, which is how we will refer to them.  Two issues with this approach are that estimating the conditional expectation is challenging (Section~\ref{sec:conditional_shapley}), and that the resulting explanations will spread credit among correlated features even if the model does not directly use all of them, which may not be
% is not always
desirable (Section~\ref{sec:which_shapley_value_is_better}).

An alternative approach is to use the \textit{marginal distribution} when sampling replacement values.
% for the removed features.
That is, we ignore the values for the observed features $x_S^e$ and sample replacement values according to
% directly using
$x_{\bar S} \sim p(x_{\bar S})$. As in the previous case, the coalitional game is defined as the expectation of the prediction across this distribution.
% An alternative approach is to use a marginal distribution: the probability distribution of a subset of the features.  As before, we set present features in $S$ to be from the explicand.  However, we instead use the marginal distribution of absent features $D\setminus S$ to sample their values, irrespective of the present features (off-manifold).  The coalitional game for subset $S$ is the average of the model's prediction based on these samples.  
This approach is equivalent to averaging over baseline Shapley values with baselines drawn from the data distribution $p(x)$~\cite{sundararajan2020many}.
It also has an interpretation based in causal interventions on the feature values, but not interventions on the real-world values the features represent, interventions on the feature values in the computer going into the machine learning model. This is equivalent to assuming a flat causal graph 
% importantly this interpretation it is only equivalent to a truly causal approach~\cite{heskes2020causal} under the assumption
(i.e., a causal graph with no causal links among features)~\cite{janzing2020feature, heskes2020causal}.  The latter
% This
interpretation has led to the name interventional Shapley values, but to avoid ambiguity we opt for the name \textit{marginal Shapley values}~\cite{heskes2020causal}. 
% It is also connected to a broader method named causal Shapley values~\cite{heskes2020causal}.  In particular, this approach is the same as causal Shapley values under the assumption of a flat causal graph, where the only edges in the graph point from each feature to the outcome~\cite{janzing2020feature}.  This interpretation has been called interventional Shapley values~\cite{chen2020true}; however, we will refer to them as \textit{marginal Shapley values}~\cite{heskes2020causal}.

% An alternative approach is to use the \textit{marginal expectation} from causal inference: $v(S)=\mathbb{E}_\mathcal{D}[f(x)|\text{do}(x_{S})]$.  The \textit{do} notation is causal inference's \textit{do}-operator~\cite{pearl2009causality}.  In words, we break the dependence between the features in $S$ and the remaining features, which is analogous to \textit{intervening} on the remaining features~\cite{janzing2020feature}.  There are a few names for Shapley values based on this coalitional game: \textit{interventional shapley values}~\cite{chen2020true}, \textit{marginal Shapley values}~\cite{heskes2020causal}, \textit{random baseline Shapley value}~\cite{sundararajan2020many}.  Additionally, this is exactly the assumption made by \href{https://shap.readthedocs.io/en/latest/#shap.KernelExplainer"}{Kernel Explainer}~\cite{lundberg2017unified} and \href{https://shap.readthedocs.io/en/latest/#shap.SamplingExplainer}{Sampling Explainer} from the SHAP package.  We will refer to these as \textit{marginal Shapley values}.

The conditional and marginal approaches are by far the most common feature removal approaches in practice. Two other formulations based on random sampling are (1)~the uniform approach, where absent features are drawn from a uniform distribution covering the feature range, and (2)~the product of marginals approach, where absent features are drawn from their individual marginal distributions (which assumes independence between all absent features)~\cite{datta2016algorithmic, merrick2020explanation}.  However, these distributions make a strong assumption of independence between all features, which may be why marginal Shapley values, which make a milder assumption of independence between the observed and unobserved features,
% less restrictive assumption,
are more commonly used.
In addition, there are several other approaches for handling absent features in Shapley value-like explanations, but these can often be interpreted as approximations of the aforementioned approaches~\cite{covert2021explaining}. 
% For simplicity, we visualize the empirical versions of each of the
We visualized the three main
removal approaches in Figure~\ref{fig:absent_features}d, where, for simplicity, we show empirical versions that
% the empirical games
use a finite set of baselines (e.g., a training data set) to compute each expectation~\cite{sundararajan2020many}.

% For the uniform, product of marginals, and marginal approaches, the empirical distribution works quite well.  However, for the conditional approach, having many features or continuous features causes the empirical conditional expectations to produce poor estimates.

\subsection{Shapley value explanations for linear models}
\label{sec:shapley_linear_examples}

% \textcolor{blue}{Ian: it may be worth rethinking the next couple paragraphs. This part puts a lot of emphasis on linear models (which are probably the least important use-case for SHAP), and it's complicated enough that it may not provide intuition for readers new to the topic.}

To highlight intuitive differences between baseline, marginal, and conditional Shapley values, we consider the case of linear models where Shapley value explanations are easier to compute and compare (Figure~\ref{fig:linear_example}).  In the following examples, we consider different linear models, data distributions, and feature encodings to call attention to properties of these three types of Shapley value explanations.

Baseline Shapley values are simple and can be intuitive, but choosing an appropriate baseline is difficult.  
One common baseline is a sample with all zero feature values.  
However, we show that an all-zeros baseline can produce
counterintuitive attribution values because the meaning of zero can be arbitrary (Figure~\ref{fig:linear_example}a).  In particular, we consider a case with equivalent
% show the equivalent
% same
models and explicand, but where the gender feature is encoded such that male is $0$ ($x_3$) or male is $1$ ($x_3'$).  In the first case, the baseline Shapley value for $x_3$ is zero (Figure~\ref{fig:linear_example}a left), signaling that being male does not impact the prediction; however, in the second case, the baseline Shapley value for $x_3'$ is $-10$ (Figure~\ref{fig:linear_example}a right), suggesting that being male leads to lower predictions.  These differing explanations are perhaps counterintuitive because the model and explicand are exactly equivalent, but
% the same.
% However,
the discrepancy arises because the meaning of zero is often arbitrary.
% , and using
% % so selecting
% it as a baseline can result in misleading attributions.

As an alternative to the all-zeros baseline,
% Alternatively,
the mean baseline is arguably a reasonable choice
% of baseline
for linear models.  When using this baseline value, as in Figure~\ref{fig:linear_example}b, the baseline Shapley value is zero for height and weight because the explicand's height and weight are equal to their average values.
% exactly average.
In addition, for a mean baseline, the baseline Shapley value is the same regardless of how we encode gender.  Although the mean baseline can work well for linear models (Section~\ref{sec:model_specific_approaches}), it can be unappealing in other cases for two reasons.  First, the mean of discrete features generally does not
% may not
have a natural interpretation.  For instance, the mean of the gender variable
% baseline of gender
is half female and half male, a value that is never encountered in the dataset.  This issue is compounded for non-ordinal discrete
% features
and categorical features with more than two possible values.  Second, it may be impossible for any single baseline to represent the absence of feature information.
% feature removal.
For example, in images, removing features with a
% single
baseline cannot give credit to pixels that match the baseline~\cite{covert2021explaining,sturmfels2020visualizing,chen2021explaining}; for a mean baseline, this means that regions of images that resemble the mean will be biased towards lower importance.
% to be less important.  

% Instead of
Rather than baseline Shapley values, we may instead prefer to use marginal and conditional Shapley values, which we compare in Figure~\ref{fig:linear_example}c.  In this example, we compute Shapley value explanations for the same explicand and baseline, but with different models and data distributions (multivariate Gaussians with different covariances).  We generate data in
% one of
two ways: independent (zero covariance between features) or dependent (high covariance between features $x_2$ and $x_3$).  In addition, we consider two models: full (all coefficients are non-zero) or partial ($\beta_3$ is zero).

Comparing the independent full model case to the dependent full model case, we can see that conditional Shapley values split credit between correlated features.  This behavior may be desirable if we want to detect whether a model is relying on a protected class through correlated features.  However, spreading credit can feel unnatural in the dependent partial model case, where the conditional Shapley value for feature $x_3$ ($\phi^c_3$) is as high  as the conditional Shapley value for feature $x_2$ ($\phi^c_2$) even though feature $x_3$ is not explicitly used by the model ($\beta_3=0$).  In particular, a common intuition is that features not algebraically used by the model should have zero attribution\footnote{This intuition is described by \textcite{sundararajan2020many} as the \textit{dummy axiom}~\cite{sundararajan2020many}. Notably, their axiom
% where the axiom
is defined
% based on features that are not used by the model.
relative to the model,
% rather than the coalitional game, 
whereas 
the game theory literature has an existing
% already defined a
dummy axiom defined relative to
% that references
the coalitional game~\cite{monderer2002variations}.
% and
Shapley value explanations
% will
always satisfy the original dummy axiom,
% with respect to their coalitional game,
as well as all other Shapley value axioms defined in terms of the coalitional game~\cite{shapley1953value, monderer2002variations}.
% The overloading of such terminology has led to confusion in the literature, because Shapley value explanations do not necessarily satisfy newly proposed axioms defined with respect to the machine learning model.
}
% \footnote{Note that this intuition is described as a dummy axiom introduced in \textcite{sundararajan2020many}, defined for machine learning models, rather than for coalitional games. However, in game theory literature there is a dummy axiom which references the coalitional games.  Shapley value explanations will always satisfy dummy axioms with respect to their coalitional games, but will not necessarily satisfy the dummy axiom with respect to their machine learning models.}
\cite{sundararajan2020many}.  One concrete example is within a mortality prediction setting (NHANES; Appendix Section~\ref{sec:data:nhanes}), where \textcite{chen2020true}~\cite{chen2020true} show that for a model that does not explicitly use body mass index (BMI) as a feature, conditional Shapley values still give high importance to BMI due to correlations with other influential features such as arm circumference and systolic blood pressure.

\subsection{Tradeoffs between removal approaches}
\label{sec:which_shapley_value_is_better}

Given the many ways to formulate the coalitional game, or
% equivalently
to handle absent features,
a natural question is \textit{which Shapley value explanation is preferred}?  This question is frequently debated in Shapley value literature, with some papers defending marginal Shapley values~\cite{sundararajan2020many,janzing2020feature}, others advocating for conditional Shapley values~\cite{ancona2019explaining,aas2019explaining,covert2021explaining}, and still others arguing for causal solutions~\cite{heskes2020causal}.  Before discussing differences, one way the approaches are alike is that each Shapley value explanation variant always satisfies the same axioms for its corresponding coalitional game, although the interpretation of the axioms can differ; this point has been discussed in prior work~\cite{sundararajan2020many}, but it is important to avoid conflating axioms defined relative to the coalitional game and relative to the model.
Below, we discuss tradeoffs between the two most popular approaches, marginal and conditional Shapley values, because these are most commonly implemented in public repositories and discussed in the literature.

% This multiplicity is the basis of criticisms of Shapley value explanations~\cite{kumar2020problems}, but this issue is in fact not unique to Shapley values:
% % explanations:
% it represents
% % and is
% a fundamental challenge in handling correlated features and is encountered by many other model explanation methods~\cite{covert2021explaining}.   
% explanations in the presence of dependent features is a broader challenge underlying more than just Shapley value explanations. 

% \hugh{Off-manifold and on-manifold paragraph here.}
% One weakness of this approach is that it evaluates
% will likely evaluate
% the model with potentially unrealistic, \textit{off-manifold} samples~\cite{frye2020shapley},
% on unrealistic (off-manifold) samples,
% for which the model's behavior may be unpredictable and uninformative.

% We will first describe them from the perspective of linear models, where it is easier to compute conditional expectations and describe some intuitive difference between both approaches.

% We first describe marginal Shapley values and conditional Shapley from the perspective of linear models, because although computing conditional Shapley is hard in general, it is easy for linear models with normally distributed data.

As we have seen with linear models, conditional Shapley values tend to spread credit between correlated features, which can surface hidden dependencies~\cite{frye2020shapley},
% (e.g., when a model uses zip code as a proxy for race), 
whereas marginal Shapley values yield attributions that are a description of the model's functional form~\cite{chen2020true}.  This discrepancy arises from the distributional assumptions, where conditioning on a feature implicitly introduces information about all correlated features, thereby leading groups of correlated features to share credit (Figure~\ref{fig:absent_features} conditional).  For example, if the feature \textit{weight} is introduced when \textit{BMI} is absent, then conditional Shapley values will only consider values of \textit{BMI} that make sense given the known value of \textit{weight} (i.e., ``on-manifold'' values); as a consequence, if the model depends on \textit{BMI} but not \textit{weight}, we would still observe that introducing \textit{weight} affects the conditional expectation of the model output.
In contrast, although marginal Shapley values perturb the data in less realistic ways (``off-manifold''), they are able to distinguish between correlated variables and identify whether the model functionally depends on \textit{BMI} or \textit{weight}, which is useful for model debugging~\cite{lundberg2020local}.

% Intuitively, marginal Shapley values can generate samples which are unrealistic combinations of present and absent features through a marginal distribution.  These samples are unrealistic because they fall outside of the data manifold (i.e., the space along which real world samples exist).  Evaluating the model for these ``off-manifold'' samples can be problematic because the model's behavior is less well-defined off-manifold~\cite{frye2020shapley}.  In contrast, conditional Shapley values aim to generate realistic values for absent features based on present features through a conditional distribution (``on-manifold'').

Having two popular types of Shapley value explanations has been cited as a weakness~\cite{kumar2020problems}, but this issue is not unique to Shapley values; it is encountered by a large number of model explanation methods~\cite{covert2021explaining}, and it is fundamental to understanding feature importance with correlated or statistically dependent features.  For example, with linear models, there are similar issues with handling correlation, where multicollinearity can result in different coefficients with equivalent accuracy.  One solution to handle multicollinearity for linear models is to utilize appropriate regularization (e.g., ridge regression)~\cite{chen2020true}, otherwise credit (coefficients) can be split among correlated features somewhat arbitrarily.  
In the model explanation context, correlated features represent a similar challenge, and the multiple ways of handling absent features can be understood as different approaches to disentangle credit for correlated features.

Another solution to address correlated features is to incorporate causal knowledge.  Causal inference approaches typically assume knowledge of an underlying causal graph (i.e., a directed graph where edges indicate that one feature causes another) from which correlations between features arise.  There are a number of Shapley value explanation methods that leverage an underlying causal graph, such as causal Shapley values~\cite{heskes2020causal}, asymmetric Shapley values~\cite{frye2020asymmetric}, and Shapley Flow~\cite{wang2021shapley}.  The major drawback of these approaches is that they assume prior knowledge of causal graphs that are unknown in the vast majority of applications.  For this reason, conditional and marginal Shapley values represent more viable options in many practical situations.
% we will instead focus on describing approaches to estimate marginal and conditional Shapley values.

In this paper, we advocate for marginal and conditional Shapley values because they are more practical than causal Shapley values, and they avoid the problematic choice of a fixed baseline as in baseline Shapley values.
% are preferable to baseline Shapley values because the correct choice of baseline is often unclear.
In addition, they cover two of the most common use-cases for Shapley value explanations and model interpretation in general: (1)~understanding a model's informational dependencies, and (2)~understanding the model's functional form.  An important final distinction between marginal and conditional Shapley values is the ease of estimation. As we discuss next in Section~\ref{sec:shapley_explanation_algorithms}, marginal Shapley values turn out to be much simpler to estimate than conditional Shapley values.
% Marginal Shapley values turn out to be much easier to estimate than conditional Shapley values, as we discuss in Section~\ref{sec:shapley_explanation_algorithms}.
% In this regard, marginal Shapley values are much easier to estimate than
% % in comparison to
% conditional Shapley values (Section~\ref{sec:shapley_explanation_type}).  In fact, for linear and tree models, it is possible to exactly compute marginal Shapley values values in polynomial time, whereas estimating conditional Shapley values is more computationally costly and potentially requires imprecise approximations~\cite{chen2020true,lundberg2020local} (Section~\ref{sec:model_specific_approaches}).

\section{Algorithms to estimate Shapley value explanations}
\label{sec:shapley_explanation_algorithms}

% As previously mentioned, there are two main challenges when estimating
% % factors of complexity in the estimation of
% Shapley value explanations: (1)~choosing the feature removal approach (or equivalently, defining the coalitional game), and (2)~tractably calculating Shapley values despite the exponential number of terms in the summation.  Here, we describe a variety of algorithmic approaches designed to address these challenges.

Here, we describe algorithmic approaches to address the two main challenges for generating Shapley value explanations: (1)~removing features to estimate the coalitional game, and (2)~tractably calculating Shapley values despite their exponential complexity.
% the exponential number of terms in the summation.

\begin{table}[!ht]
% \begin{tabular}{llllll}
\centering
\resizebox{\textwidth}{!}{%

\begin{tabular}{@{}*{1}{L{\dimexpr0.35\textwidth-2\tabcolsep\relax}}*{6}{L{\dimexpr0.13\textwidth-2\tabcolsep\relax}}@{}}
\toprule
& \multicolumn{3}{c}{Factors of complexity} &
\multicolumn{3}{c}{Properties} \\ \cmidrule(r{4pt}){2-4} \cmidrule(l){5-7}
Method                                                       & Estimation strategy & Removal approach & Removal variant & Model-agnostic & Bias-free         & Variance-free    \\ \midrule
ApproSemivalue~\cite{castro2009polynomial}                   & SV                  & None             & Exact            & Yes            & Yes               & No               \\
L-Shapley~\cite{chen2018shapley}                             & SV                  & Marginal         & Empirical       & Yes            & No                & No$^{\clubsuit}$ \\
C-Shapley~\cite{chen2018shapley}                             & SV                  & Marginal         & Empirical       & Yes            & No                & No$^{\clubsuit}$ \\
ApproShapley~\cite{castro2009polynomial}                     & RO                  & None             & Exact            & Yes            & Yes               & No               \\
IME~\cite{strumbelj2010efficient}                            & RO                  & Marginal         & Empirical       & Yes            & Yes               & No               \\
CES~\cite{sundararajan2020many}                              & RO                  & Conditional      & Empirical       & Yes            & No                & No               \\
Shapley cohort refinement~\cite{mase2019explaining}          & RO                  & Conditional      & Empirical*      & Yes            & No                & No               \\
Generative model~\cite{frye2020shapley}                      & RO                  & Conditional      & Generative      & Yes            & No                & No               \\
Surrogate model~\cite{frye2020shapley}                       & RO                  & Conditional      & Surrogate       & Yes            & No                & No               \\
Multilinear extension sampling~\cite{okhrati2021multilinear} & ME                  & Marginal         & Empirical       & Yes            & Yes$^\diamondsuit$               & No               \\
SGD-Shapley~\cite{simon2020projected}                  & WLS                 & Baseline         & Exact           & Yes            & No$^{\heartsuit}$ & No               \\
KernelSHAP~\cite{lundberg2017unified,covert2020improving}   & WLS                 & Marginal         & Empirical       & Yes            & Yes$^{\spadesuit}$              & No               \\
Parametric KernelSHAP~\cite{aas2019explaining}              & WLS                 & Conditional      & Parametric      & Yes            & No                & No               \\
Nonparameteric KernelSHAP~\cite{aas2019explaining}          & WLS                 & Conditional      & Empirical*      & Yes            & No                & No               \\
FastSHAP~\cite{jethani2021fastshap}                          & WLS                 & Conditional      & Surrogate       & Yes            & No                & No               \\ \midrule
LinearSHAP~\cite{chen2020true}                              & Linear              & Marginal         & Empirical       & No             & Yes               & Yes              \\
Correlated LinearSHAP~\cite{chen2020true}                   & Linear              & Conditional      & Parametric      & No             & No                & No               \\
Interventional TreeSHAP~\cite{lundberg2020local}            & Tree                & Marginal         & Empirical       & No             & Yes               & Yes              \\
Path-dependent TreeSHAP~\cite{lundberg2020local}            & Tree                & Conditional      & Empirical*      & No             & No                & Yes              \\
DeepLIFT~\cite{shrikumar2017learning}                        & Deep                & Baseline         & Exact           & No             & No                & Yes              \\
DeepSHAP~\cite{lundberg2017unified}                         & Deep                & Marginal         & Empirical       & No             & No                & Yes              \\
DASP~\cite{ancona2019explaining}                             & Deep                & Baseline         & Exact           & No             & No                & No$^{\clubsuit}$ \\
Shallow ShapNet~\cite{wang2020shapley}                       & Deep                & Baseline         & Exact           & No             & Yes               & Yes              \\
Deep ShapNet~\cite{wang2020shapley}                          & Deep                & Baseline         & Exact           & No             & No                & Yes              \\
\bottomrule
\end{tabular}}
\caption{Methods to estimate Shapley value explanations.  We order approaches based on whether or not they are model-agnostic.  Then, there are two factors of complexity.  The first is the estimation strategy to handle the exponential complexity of Shapley values.   For the model-agnostic approaches, the strategies include semivalue (SV), random order value (RO), multilinear extension (ME), and least squares value (LS).  Note that the model-agnostic estimation strategies can generally be adapted to apply for any removal approach.  For model-specific approaches, the strategies differ for linear, tree, and deep models.  Then, the second factor of complexity is the feature removal approach which determines the type of Shapley value explanation (Section~\ref{sec:shapley_explanation_type}).  ``Any'' denotes that it was introduced in game theory, and not for the sake of explaining a machine learning model.  Then, we describe the specific removal variant employed by each algorithm.  Baseline Shapley values are always computed exactly (Section~\ref{sec:baseline_shapley}), marginal Shapley values are always estimated empirically (Section~\ref{sec:marginal_shapley}), and conditional Shapley values have a variety of estimation procedures (Section~\ref{sec:conditional_shapley}).  *These empirical estimates also involve defining a similarity metric.  Finally, we report whether approaches are bias-free and/or variance-free.  $^\diamondsuit$Multilinear extension sampling is unbiased when sampling $q$ uniformly.  However, it is more common to use the trapezoid rule to determine $q$ which improves convergence, but can lead to higher bias empirically at smaller numbers of subsets (Appendix Tables \ref{tab:diabetes_bias}, \ref{tab:nhanes_bias}, \ref{tab:blog_bias}).  $^{\heartsuit}$SGD-Shapley is consistent, but based on our empirical analysis it has high bias relative to other approaches (Appendix Tables \ref{tab:diabetes_bias}, \ref{tab:nhanes_bias}, \ref{tab:blog_bias}).  $^{\spadesuit}$One version of KernelSHAP has been proven to be bias-free and the original version is asymptotically unbiased~\cite{williamson2020efficient}, although empirically it also appears to be unbiased for moderate numbers of samples~\cite{covert2020improving}.
% empirically appears to be bias-free~\cite{covert2020improving}.
$^{\clubsuit}$These approaches can be deterministic with a polynomial number of model evaluations, but are often run with fewer evaluations for computational speed.}
\label{tab:methods}
\end{table}

\subsection{Feature removal approaches}
\label{sec:shapley_explanation_type}

Previously we introduced three main feature removal approaches and discussed tradeoffs between them.  In this section, we
% will
discuss how to calculate the coalitional games that correspond to
% which in turn produce each of
the most popular variants of Shapley value explanations: baseline Shapley values, marginal Shapley values, and conditional Shapley values.

\subsubsection{Baseline Shapley values}
\label{sec:baseline_shapley}

The coalitional game for baseline Shapley values is defined as
% as follows:
\begin{equation}
v(S) = f(x^e_S, x^b_{\bar{S}}), \label{eq:baseline_shapley_game}
\end{equation}

\noindent where $f(x^e_S, x^b_{\bar{S}})$ denotes evaluating $f$ on a hybrid sample where present features are taken from the explicand $x^e$ and absent features are taken from the baseline $x^b$.  To compute the value of this coalitional game, we can
% algorithms
simply create a hybrid sample and then return the model's prediction for that sample.  It is possible to exactly compute this coalitional game, unlike the remaining approaches.  The only parameter is the choice of baseline, which can be a somewhat arbitrary decision. 
% a controversial decision.

\subsubsection{Marginal Shapley values}
\label{sec:marginal_shapley}

For marginal Shapley values, the coalitional game is the marginal expectation of the model output,
\begin{equation}
% v(S) = \mathbb{E}_{X_{\bar{S}}}[f(x^e_S, X_{\bar{S}})],
v(S) = \mathbb{E}_{p(x_{\bar{S}})}[f(x^e_S, x_{\bar{S}})],
\end{equation}

\noindent where $x_{\bar{S}}$ is treated as a random variable representing the missing features and we take the expectation over the marginal distribution $p(x_{\bar S})$ for these missing features.

% The most
A natural approach to compute the marginal expectation is to leverage the training or test data
% data drawn from some distribution $D$ (training or test data)
to calculate an empirical estimate. A standard assumption in machine learning is that the data are independent draws from the data distribution $p(x)$, so we can designate a set of observed samples $E$ as an empirical distribution
% For the marginal expectation, the dependence between features in the set $S$ and the remaining features is broken,
% so we can draw baseline samples from an empirical distribution $\hat{D}$ (i.e., a set of observed samples with equal probability)
and use their values for the absent features (Figure~\ref{fig:absent_features}d Marginal):
\begin{equation}
% v(S) = \frac{1}{|\hat{D}|}\sum_{x^b\in \hat{D}} f(x^e_S, x^b_{\bar{S}}) \label{eq:marginal_shapley_empirical_game}
v(S) = \frac{1}{|E|}\sum_{x^b\in E} f(x^e_S, x^b_{\bar{S}}). \label{eq:marginal_shapley_empirical_game}
\end{equation}

From Eq.~\ref{eq:marginal_shapley_empirical_game}, it is clear that the empirical marginal expectation is the average over the coalitional games for baseline Shapley values with many baselines (Eq. \ref{eq:baseline_shapley_game}). As a consequence, 
% However, perhaps more surprisingly,
marginal Shapley values
% \footnote{Based on the empirical marginal expectation.}
are also the average over many baseline Shapley values~\cite{chen2021explaining}.  Due to this, some algorithms estimate marginal Shapley values by first estimating baseline Shapley values for many baselines and then averaging them~\cite{lundberg2020local,chen2021explaining}.  Note that marginal Shapley values based on empirical estimates are unbiased if the baselines are drawn i.i.d. from the baseline distribution (e.g., a random subset of rows from the dataset).  As such, empirical estimates are considered a reliable
way to approximate the true marginal expectation.

The empirical distribution can be the entire training dataset, but in practice it is often a moderate number of samples from the training or test data~\cite{sundararajan2020many, lundberg2020local}.  The primary parameter is the number of baseline samples and how to choose them.  If a large number of baselines is chosen, they can safely be chosen uniformly at random; however, when using a smaller number of samples, approaches based on k-means clustering can be used to ensure better coverage of the data distribution.  This empirical approach also applies to other coalitional games such as the uniform and product of marginals, which are similarly easy to estimate~\cite{merrick2020explanation}.

\subsubsection{Conditional Shapley values}
\label{sec:conditional_shapley}

For conditional Shapley values, the coalitional game is the conditional expectation of the model output,
\begin{equation}
% v(S) = \mathbb{E}_{X_{\bar{S}} \mid X_{S}=x^e_S}[f(x^e_S, X_{\bar{S}})],
v(S) = \mathbb{E}_{p(x_{\bar{S}} \mid x^e_S)}[f(x^e_S, x_{\bar{S}})],
\end{equation}

\noindent where $x_{\bar S}$
% $X_{\bar{S}}$
is considered a random variable representing the missing features, and we take the expectation over the conditional distribution $p(x_{\bar S} \mid x^e_S)$ of these missing features given the known features $x^e_S$ from the explicand.

Computing conditional Shapley values is more difficult
% much harder
because the required conditional distributions
% conditional expectations
are not readily available from the training data.  We can empirically estimate
% Empirically estimating
conditional expectations by averaging model predictions from samples that match the explicand's present features (Figure~\ref{fig:absent_features}d conditional), and this exactly estimates
% would exactly estimate
the conditional expectations as the number of baseline samples goes to infinity.  However, this empirical estimate does not work well in practice: the number of matching rows may be too low in the presence of continuous features or a large number of features, leading to inaccurate and unreliable estimates
% because it produces inaccurate estimates in the presence of continuous features or when there are large numbers of features 
\cite{sundararajan2020many}.
% In particular, the number of matching rows may be too low to be useful.
For instance, if one conditions on a height of $5.879$ feet, there are likely very few individuals with that exact height, so the empirical conditional expectation will average over very few samples' predictions, or potentially
% even
just the single prediction from the explicand itself.
% be the average of a very few, or even zero, samples' predictions.

One natural solution is to approximate the conditional expectation based on similar feature values rather than exact matches~\cite{mase2019explaining, sundararajan2020many,aas2019explaining}.  For instance, rather than condition on baselines that are $5.879$ feet tall, we can condition on baselines that are between $5.879\pm 0.025$ feet tall.  This approach requires a definition of
% way to define
similarity, which is not obvious and may be an undesirable prerequisite for an explanation method.  Furthermore, these approaches do not fully solve the curse of dimensionality, and conditioning on many features can still lead to inaccurate estimates of the conditional expectation.

% \hugh{Figure comparing the convergence under different modeling assumptions for normal data.  Could incorporate real data here, but I think it would be very hard to express the differences when we don't know the data generating mechanism.  One possibility is bulk RNA-seq data which is roughly normal when log normalized.}

Instead of empirical estimates, a number of approaches based on fitting models have been proposed to estimate the conditional expectations.  Many of these have been identified in the broader context of removal-based explanations~\cite{covert2021explaining}, but we reiterate them here, summarizing practical strengths and weaknesses:

% In response to these issues with empirical estimates, a better \textit{target coalitional game is the unknown, underlying conditional expectation}.  There are two broad categories of approaches that attempt to learn this expectation: (1)~parametric models and (2)~deep models:

\begin{itemize}
    \item \textbf{Parametric assumptions.}  \textcite{chen2020true} and \textcite{aas2019explaining} assume Gaussian or Gaussian-copula distributions.  Conditional expectations for Gaussian random variables have closed-form solutions and are computationally efficient once the joint distribution's parameters have been estimated, but these approaches can have large bias if the parametric assumptions are incorrect.
    \item \textbf{Generative model.}  \textcite{frye2020shapley} use a conditional generative model
    % variational autoencoder
    to learn the conditional distributions given every subset of features.
    % $S$.  
    % This approaches requires that the generative model accurately captures an exponential number of distributions, which is a challenging modeling task, particularly given a large number of features.
    % The key assumption of this approach is that the variational autoencoder, which becomes a new parameter, accurately captures each of the exponential number of conditional distributions.  
    The generative model provides samples from approximate conditional distributions, and with these
    % for which
    we can average model predictions to estimate the conditional expectation.
    % Then, we can estimate the conditional expectation by averaging the model prediction for samples drawn according to the generative model.  
    In general, this approach is more flexible than simple parametric assumptions, but it has variance due to the stochastic nature of training deep generative models, and it is difficult to assess whether the generative model accurately approximates the exponential number of conditional distributions.
    % \hugh{Still biased with infinite training time/data b/c of VAE assumptions?} 
    % Assumption, runtime, and bias/variance, evaluation.  \hugh{Fundamental limitations based on data availability?  Kind of true of marginal too?}
    \item \textbf{Surrogate model.} \textcite{frye2020shapley} use a surrogate model to learn the conditional expectation of the original model given every subset of features.
    % $S$.  
    The surrogate model is trained to match the original model's predictions with arbitrarily held-out features, and doing so has been shown to directly approximate the conditional expectation, both for regression and classification models~\cite{covert2021explaining}.  This approach is as flexible as the generative model, but it has several practical advantages: it is simpler to train, it requires only one model evaluation to estimate the conditional expectation, and it has been shown to provide a more accurate estimate in practice~\cite{frye2020shapley}.
    % As with the generative model approach, the surrogate model becomes a new parameter and introduces variance.  
    \item \textbf{Missingness during training.}  \textcite{covert2021explaining} describe an approach for directly estimating the conditional expectation by training the original model to accommodate missing features.  Unlike the previous approaches, this approach cannot be applied post-hoc with arbitrary models because it requires modifying the training process.
    % Furthermore its ability to approximate the conditional expectation is based on an assumption of model optimality.
    % This approach assumes a specific loss formulation and that the model is optimal.  Then, it is equivalent to the conditional expectation.  One downside of this approach is that it requires modifying the model and the training process which means that it is not model-agnostic: it can only be used for models that can accommodate absent features.
    \item \textbf{Separate models.} \textcite{lipovetsky2001analysis}, \textcite{vstrumbelj2009explaining}, and \textcite{williamson2020efficient} directly estimate the conditional expectation given a subset of features as the output of a model trained with that feature subset.
    % subset of features.
    If every model is optimal (e.g., the Bayes classifier), then the conditional expectation estimate is exact~\cite{covert2021explaining}. In practice, however,
    % in practice
    the various models will be sub-optimal and unrelated to the original one, making it unsatisfying to view it as an explanation for the original model trained on all features. Furthermore, the computational demands of training models with many
    % various
    feature subsets is significant, particularly for non-linear models such as tree ensembles and neural networks.
    % on the subset of features $S$.  
    % If every model is optimal, then the estimate is exact.  Unfortunately, it is generally impossible to obtain optimal models; instead, each subset model is a machine learning model (e.g., linear model, tree model, deep model).  As a practical consequence of this, it can be unsatisfying to view this approach as an explanation for a specific model.  For example, if one is trying to explain a poorly optimized tree model, then this approach may simply use highly optimized tree models that are not representative of the original model to be explained.  Another practical downside of this approach is that it is extremely expensive to train models for every possible subset.
\end{itemize}

% \hugh{Evaluation, how do we know we have good estimates?}

As we have just shown, there are a wide variety of approaches to model conditional distributions or directly estimate the conditional expectations.  These approaches will generally be biased, or inexact, because the coalitional game we require is based on the true
% truly want is the
underlying conditional expectation.  Compounding this, it is difficult to quantify the approximation quality
% bias
because the conditional expectations are unknown, except in very simple cases (e.g., synthetic multivariate Gaussian data).

Of these approaches, the empirical approach produces poor estimates, parametric approaches
% assumptions
require strong assumptions, missingness during training is not model-agnostic, and separate models is not exactly an explanation of the original model.  Instead, we believe approaches based on a generative model or a surrogate model are more promising.  These approaches are more flexible, but both require fitting an additional deep model.  To assess these deep models, \textcite{frye2020shapley} propose two reasonable metrics based on mean squared error of the model's output to evaluate the generative and surrogate model approaches.   Future work may include identifying robust architectures/hyperparameter optimization for surrogate and generative models, analyzing how conditional Shapley value estimates change for non-optimal surrogate and generative models, and
% perhaps
evaluating bias
% approximation quality
in conditional Shapley value estimates for data with known conditional distributions.

Some of the approaches we discussed approximate the intermediate conditional distributions (empirical, parametric assumptions, generative model) whereas others directly approximate conditional expectations (surrogate model, missingness during training, separate models).  It is worth noting that approaches based on modeling conditional distributions are independent of the particular model $f$.  This suggests that if a researcher fits a high-quality generative model to a popular dataset, then any subsequent researchers can re-use this generative model to estimate conditional Shapley values for their own predictive models.  However, even if fit properly, approaches based on modeling conditional distributions
% conditional distribution approaches
may be more computationally expensive, because they 
% generally
require evaluating the model with
% on
many generated samples to estimate the conditional expectation.  As such, the surrogate model approach may be more effective than the generative model approach in practice~\cite{frye2020shapley}, and it has been used successfully in recent work~\cite{jethani2021fastshap, covert2022learning}.

% Although approaches that rely on parametric assumptions may have variance-free estimates of the conditional expectations,  conditional Shapley values based on parametric models will generally have larger bias, because they are less flexible.  In comparison, the generative model approach also learns conditional distributions, but will be

% There are a variety of methods to estimate conditional distributions.  These approaches will be biased, because the coalitional game we truly want is the underlying conditional expectation which is difficult to estimate.  Some of them will also depend on fitting models (deep models approaches), which can be a stochastic process that will introduce variance as well.  

In summary, in order to compute conditional Shapley values, there are \textit{two primary parameters}: (1)~the approach to model the conditional expectation, for which there are several choices.
% of which there are many. 
Furthermore, within the approaches that rely on deep models (generative model and surrogate model), the training and architecture of the deep model becomes an important
% a critical
yet complex dependency.  (2)~The baseline set used to estimate the conditional distribution or model the conditional expectation, because each approach requires a set of baselines (e.g., the training dataset) to learn dependencies between features.  Different sets of baselines can lead to different scientific questions~\cite{merrick2020explanation}.
For instance, using baselines drawn from older male subpopulations, we can ask "\textit{why does an older male individual have a mortality risk of X\% relative to
the subpopulation of older males?}"~\cite{chen2021explaining}.
% an older male subpopulation?
% (3)~Any additional features to include in the coalitional game, because, as discussed in Section~\ref{sec:which_shapley_value_is_better}, conditional Shapley value explanations spread credit between correlated features, even if they are not used by the model.

\subsection{Tractable estimation strategies}
\label{sec:complexity}

Calculating Shapley values is, in the general case, an NP-hard problem
% Due to how they are defined,
% the computational complexity to calculate Shapley values is NP-hard in general 
\cite{deng1994complexity,faigle1992shapley}.  Intuitively, a brute force calculation based on Eq.~\ref{eq:shapley_definition} has
% Similarly, the brute force computation of feature attributions based on Shapley values has an 
exponential complexity in the number of features because it involves
% requires
evaluating the model with all possible feature subsets.
Given the long history of Shapley values, there is naturally considerable
% a considerable amount of
research into their calculation.
% computation.
% are naturally many pre-existing estimation algorithms from game theory.
Within the game theory literature, two types of estimation strategies have emerged~\cite{castro2017improving,fatima2008linear,illes2019estimation}: (1)~approximation-based strategies that produce unbiased
% , stochastic
Shapley value estimates
% estimators
for any game~\cite{castro2009polynomial}, and (2)~assumption-based strategies that can produce exact results
% estimators
in polynomial time for specific types of games~\cite{megiddo1978computational,granot2002cost,castro2009polynomial}. 
% \hugh{Add citations}

% \hugh{Traditionally assumptions were more popular and approximations less so, but vice versa in ML literature}

These two strategies have also prevailed in Shapley value explanation literature.  However, because some approaches rely on both assumptions and approximations, we instead categorize the approaches as \textit{model-agnostic} or \textit{model-specific}.   \textit{Model-agnostic} estimation approaches
% approaches to estimate Shapley value explanations
make no assumptions on the model class and often rely on stochastic, sampling-based estimators
% approximations to produce stochastic estimators 
\cite{strumbelj2010efficient,lundberg2017unified,okhrati2021multilinear,jethani2021fastshap}. In contrast, \textit{model-specific} approaches rely on assumptions about
% on
the machine learning model's class to improve the speed of calculation, although sometimes at the expense of exactness~\cite{chen2020true,lundberg2020local,chen2021explaining,ancona2019explaining,wang2020shapley}. 

% Unfortunately, the assumption based approaches for traditional Shapley values often rely on assuming very specific and often times simple games.  These approaches do not generalize well to Shapley value explanations which construct games based on machine learning models.  However, 

\subsubsection{Model-agnostic approaches}
\label{sec:model_agnostic_approaches}

There are several types
% a few varieties
of model-agnostic approaches to estimate Shapley value explanations.  In general, these
% The most common strategy is to use
are approximations that sidestep evaluating the model with an exponential number of subsets by instead using
% sampling
a smaller number of subsets
% that are
chosen at random.
% by instead only evaluating a polynomial number of subsets.
% As a result,
These approaches are generally
% stochastic but consistent: that is, their results are non-deterministic, but they are guaranteed to be correct given an infinite number of samples.
unbiased but stochastic; that is, their results are non-deterministic, but they are correct in expectation.
% and tend to improve with the number of samples used,

To introduce these approaches, we describe how each approximation
% approach
% is motivated by
can be tied to
a distinct mathematical characterization of the Shapley value. We provided one such characterization in Section~\ref{sec:shapley_values} (Eq.~\ref{eq:shapley_definition}), but there are multiple equations to represent the Shapley value, and each one suggests its
% that each suggest their
own approximation approach. For simplicity, we discuss these approaches in the context of
% traditional Shapley values based on 
a game $v$
% which generalizes to Shapley value explanations.
where we ignore the choice of feature removal technique (baseline, marginal or conditional).

% There are many equivalent ways to characterize the Shapley value.
The classic Shapley value definition is as a \textbf{semivalue}~\cite{dubey1981value}, where each player's credit
% which
is a weighted average
% weighted summation
of the player's marginal contributions. For this we require a weighting function $P(S)$ that depends only on the subset's cardinality, or where $\sum_{S \subseteq D \setminus \{i\}} P(S) = 1$ for $i = 1, \ldots, d$.
% , or $P(S) = \rho(|S|)$ for some function $\rho: \{0, \ldots, d\} \mapsto \mathbb{R}$.
% for each player,
Then, the value for player $i$ is given by

\begin{equation}
% \phi_i(v)=\sum_{S\subseteq D\setminus \{i\}} P(S) \big(v(S\cup \{i\}) - v(S)\big).
\phi_i(v)=
% \frac{1}{\sum_{S \subset D \setminus \{i\}}P(S)}
% \frac{1}{c}
% \sum_{S\subseteq D\setminus \{i\}} P(S) \big(v(S\cup \{i\}) - v(S)\big),
% \frac{1}{c}
\sum_{S\subseteq D\setminus \{i\}} P(S) \big(v(S\cup \{i\}) - v(S)\big).
\end{equation}

% \noindent where $c = 1 / \sum_{S \subseteq D \setminus \{i\}} P(S)$, and $c$ is identical for $i = 1, \ldots, d$.
% \noindent where
% $C(\cdot)$ is defined such that
% $\sum_{S \subseteq D \setminus \{i\}} P(S) = 1$ for $i = 1, \ldots, d$.
 
% Based on the view of Shapley values as a semivalue
\textcite{castro2009polynomial} proposed an unbiased, stochastic estimator (ApproSemivalue) for \textit{any} semivalue (i.e., with arbitrary weighting function)
% $P(S)$)
% by sampling
that involves sampling subsets from $D \setminus \{i\}$ with probability given by
% proportional to
$P(S)$.
% according to the probability distribution
% a probability mass function, where the probability of a subset $S$ is
% given by $P(S)$.
% $\frac{P(S)}{\sum_{S\subseteq D\setminus \{i\}} P(S)}$.
% Despite its flexibility, it is uncommon to use ApproSemiValue to estimate Shapley value explanations, because it does not specify how to draw samples from the probability mass function.
% In the case of the Shapley value, we can apply the algorithm using the Shapley value's weighting function, which \textcite{castro2009polynomial} refer to as ApproShapley.
In this algorithm, each player's Shapley value is estimated one at a time, or independently.  To use ApproSemivalue to estimate Shapley values, we simply have to draw subsets according to the
distribution
% weighting function 
$P(S) = \frac{|S|!(|D|-|S|-1)!}{|D|!}$.
While apparently simple, Shapley value estimators inspired directly by the semivalue characterization are
% this approach is
uncommon in practice because
% it is not simple to sample from this $P(S)$.
sampling subsets from $P(S)$ is not straightforward.
% , but this is uncommon.  

Two related approaches are Local Shapley (L-Shapley) and Connected Shapley (C-Shapley)~\cite{chen2018shapley}.  Unlike other model-agnostic approaches,
L-Shapley and C-Shapley are designed for structured data (e.g., images) where nearby features are closely related (spatial correlation).
% (e.g., sentences and images). 
Both approaches are biased Shapley value
% are biased
estimators because they restrict the game to consider only coalitions of players within the neighborhood of
% have fewer players that must be nearby
the player being explained\footnote{Technically L-Shapley and C-Shapley are \textit{probabilistic values}~\cite{monderer2002variations}, a generalization of semivalues, because their weighting functions differ based on the spatial location of the current feature.}.  They are variance-free for sufficiently small neighborhoods, but for large neighborhoods it may still be necessary to use sampling-based approximations that
% will
introduce variance.

% For L-Shapley $P(S)=0$ for subsets containing features outside of a neighborhood around the feature being explained and for C-Shapley $P(S)=0$ for subsets that do not form a connected component within the neighborhood.  For L-Shapley the remaining weights are chosen to match Shapley value coefficients for the neighborhood and for C-Shapley the remaining weights are chosen to match the Myerson value~\cite{myerson1977graphs}.  Because both approaches use different weights than the original Shapley values they are biased estimators.  Both estimators are variance-free for a sufficiently small neighborhood; however, L-Shapley is exponential in the size of the neighborhood and C-Shapley is polynomial in the size of the neighborhood.  This means that for large enough neighborhoods, it may still be necessary to use sampling based approximations which would introduce variance in the estimators.

Next, the Shapley value can also be viewed as a \textbf{random order value}~\cite{shapley1953value,monderer2002variations}, where a player's credit is the average contribution across many
% all
possible orderings. 
% Here, $\pi$ denotes a permutation of the indices $\{1, \ldots, d\}$, $\Pi(D)$ represents the space of all such permutations, and $\pi[:i]$ denotes the set of all features before position $i$ in the permutation $\pi$.
% Here, $\pi$ denotes a permutation of the indices $\{1, \ldots, d\}$, $\Pi(D)$ represents the space of all such permutations, and $\pi[:i]$ denotes the set of all players preceding player $i$ in the permutation $\pi$.
Here, $\pi:\{1,\ldots,n\}\to\{1,\ldots,n\}$ denotes a permutation that maps from each position $j$ to the player $\pi(j)$.  Then, $\Pi(D)$ denotes the set of all possible permutations and $Pre^i(\pi)$ denotes the set of predecessors of player $i$ in the order $\pi$ (i.e., $Pre^i(\pi)=\{\pi(1),\ldots,\pi(j-1)\}$, if $i=\pi(j)$).
% $\pi[:i]$ denotes the set of all features before position $i$ in the permutation $\pi$ of the indices $\{1, \ldots, d\}$, and $\Pi(D)$ represents the space of all permutations of the players.
Then, the Shapley value's random order characterization
% random order value characterization of the Shapley value
is the following:
% an unweighted average of marginal contributions across all permutations:

\begin{equation}
\phi_i(v) = \frac{1}{|D|!} \sum_{\pi\subseteq \Pi(D)} (v(Pre^i(\pi) \cup \{i\}) - v(Pre^i(\pi))).
% \phi_i(v) = \frac{1}{|D|!} \sum_{\pi\subseteq \Pi(D)} v(\{j|\pi^{-1}(j)\leq \pi^{-1}(i)\}) - v(\{j|\pi^{-1}(j)< \pi^{-1}(i)\}).\\
% \phi_i(v) = \frac{1}{|D|!} \sum_{\pi\subseteq \Pi(D)} (v(\pi[:i+1]) - v(\pi[:i])).
\label{eq:random_order_value}
\end{equation}

% \noindent
There are two unbiased, stochastic estimation approaches based on this characterization.
% that produce unbiased, stochastic estimators.
The first approach is IME (Interactions-based Method for Explanation)~\cite{strumbelj2010efficient}, which estimates Eq.~\ref{eq:random_order_value} for each player with a fixed number of random permutations from $\Pi(D)$.
Perhaps surprisingly,
IME is analogous to ApproSemivalue, because identifying the preceding players in a random permutation can be understood as sampling from the probability distribution $P(S)$.
% sampling from the probability distribution $P(S) / c$.
% where permutations are used to draw from $P(S)$ for one player at a time.
One variant of IME improves the estimator's
% rate of
convergence by allocating more samples
% spending more time
to estimate $\phi_i(v)$ for players with high variance in their marginal contributions,
which we refer to as \textit{adaptive sampling}~\cite{vstrumbelj2014explaining}.
% (\textit{adaptive sampling})~\cite{vstrumbelj2014explaining}.

% In general, with and without the adaptive sampling, this approach will not satisfy efficiency.  To address this, SamplingExplainer in the shap github repo calculates an adjustment term based on a ridge regression that is added to the feature attribution estimates.  Doing so may introduce bias into the adaptive sampling estimator, but also guarantees efficiency of the final estimates.

% \hugh{quasi-random sampling in strumbelj paper.}

The second approach is ApproShapley, which explains all features simultaneously given a set of sampled permutations~\cite{castro2009polynomial}.  Rather than draw permutations independently for each player, this approach iteratively adds all players according to
% iterates through
each sampled permutation
% (from $\Pi(D)$)
so that all players' estimates rely on the same number of marginal contributions based on the same permutations.  There are many variants that aim to draw \textit{samples efficiently} (i.e., reduce the variance of the estimates): antithetic sampling~\cite{rubinstein2016simulation,mitchell2021sampling}, stratified sampling~\cite{maleki2015addressing,castro2017improving}, orthogonal spherical codes~\cite{mitchell2021sampling}, and more~\cite{van2018new,mitchell2021sampling,illes2019estimation}.  Of these approaches, antithetic sampling is the simplest.  After sampling a subset and evaluating its marginal contribution, antithetic sampling also evaluates the marginal contribution of the inverse of that subset ($N\setminus S$).  Recent work finds that antithetic sampling provides near-best convergence in practice compared to several more complex methods~\cite{mitchell2021sampling}.
% Recent work in this direction finds that the simple approach of antithetic sampling, which, for each subset considered, always draws the inverse of that subset, provides near-best convergence empirically~\cite{mitchell2021sampling}.
% Sobol permutations~\cite{mitchell2021sampling}, and ergodic sampling~\cite{illes2019estimation}
% One small caveat is that orthogonal spherical codes increase the complexity of sampling a permutation to $O(d^2)$, rather than $O(d)$ for antithetical sampling, where $d$ is the number of features.

The primary difference between IME and ApproShapley is that IME estimates $\phi_i(v)$ independently for each player, whereas ApproShapley estimates them simultaneously for $i = 1, \ldots, d$.  This means that IME can use adaptive sampling, which evaluates a different number of marginal contributions for each player and can greatly improve convergence when many players have low importance.  In contrast, walking through permutations as in ApproShapley is advantageous because (1)~it halves the number of evaluations of the game (which are expensive) by reusing them, and (2)~it guarantees that the efficiency axiom is satisfied (i.e., the estimated Shapley values sum to the model's prediction).  

% A final advantage is ApproShapley with antithetical sampling provides exact estimates for models that only have up to second order interaction effects with just one permutation.

% This extra complexity is likely fine for the majority of datasets including the ones evaluated in \textcite{mitchell2021sampling} (which have 30 features at most), because it is incurred once per permutation (or once every $d$ model evaluations).  However, prior to widespread adoption of these approaches, it would be important to further investigate these approaches on datasets with very large numbers of features.

The third characterization of the Shapley value is as a \textbf{least squares value}~\cite{charnes1988extremal,ruiz1998family}. In this approach, the Shapley value is viewed
% , which means that the credits are
as the solution to a weighted least squares (WLS) problem. The problem requires a weighting kernel $W(S)$, and the credits are the coefficients that minimize the following 
% problem:
objective,
% which is the solution to a weighted least squares (WLS) problem:

\begin{equation}
\phi(v) = \argmin_\beta \sum_{S\subseteq D}  W(S) (u(S) - v(S))^2,
% \phi(v) = \argmin_\beta \sum_{S\subseteq D}  W(S) (v(\emptyset) + \sum_{i \in S} \beta_i - v(S))^2 \quad \text{s.t.} \; \sum_{i \in D} \beta_i = v(D) - v(\emptyset)
\label{eq:least_squares}
\end{equation}

\noindent where $u(S)=\beta_0+\sum_{i\in S} \beta_i$ is an additive game\footnote{We present a generalized version of the least squares value, which originally involved
% by definition has
% was initially proposed with
additional constraints on the coefficients~\cite{ruiz1998family}.
% presented with an efficiency constraint and no intercept term.
}.  In order to obtain the Shapley value, we require the weighting kernel 
$W(S)=\frac{|D|-1}{{|D| \choose |S|}|S|(|D|-|S|)}$.

Based on this definition, a natural estimation approach is to sample
% the objective using
a moderate number of subsets according to $W(S)$ 
% approximate the objective function (equation \ref{eq:least_squares}),
and then solve the approximate WLS problem (Eq.~\ref{eq:least_squares}).
% stochastic estimator samples from the distribution of possible subsets according to the weighting kernel and then estimates
% and calculate the solution to the WLS problem
% in equation \ref{eq:least_squares}
% based on the approximated objective function
% these subsets
% (KernelSHAP)~\cite{lundberg2017unified, covert2020improving}.
This approach is known as \textit{KernelSHAP}~\cite{lundberg2017unified}, and its statistical properties have been studied in recent work:
% ~\cite{covert2020improving, williamson2020efficient}:
% [Comment on bias/consistency here?]
KernelSHAP is consistent and
asymptotically unbiased\footnote{\textcite{williamson2020efficient} prove this result for a global version of KernelSHAP, but it holds for the original version as well because it is an M-estimator~\cite{van2000asymptotic}.}~\cite{williamson2020efficient},
% asymptotically unbiased because it is an M-estimator~\cite{van2000asymptotic, williamson2020efficient},
and it is empirically unbiased even for a moderate number of samples~\cite{covert2020improving}.
Variants of this approach include (1)~a regularized version
% of the WLS problem
that introduces bias while reducing variance~\cite{lundberg2017unified},
% (2)~a modified version that is provably unbiased for any number of samples but has higher variance~\cite{covert2020improving},
and (2)~an antithetic version that pairs each sampled subset with its complement
% uses the complement of subsets
to improve convergence~\cite{covert2020improving}.  

Another approach that we refer to as SGD-Shapley is also based on the least squares characterization.  It samples subsets according to the weighting kernel $W(S)$, but it iteratively estimates the solution from a random initialization using projected stochastic gradient descent
% rather than solve it exactly
\cite{simon2020projected}.  Although it is possible to prove meaningful theoretical results for this strategy~\cite{simon2020projected}, our empirical evaluation shows that the KernelSHAP estimator consistently outperforms this approach in terms of
its convergence (i.e., consistently lower estimation error given an equal number of samples)
% convergence
(Appendix Figures \ref{fig:diabetes_all_error}, \ref{fig:nhanes_all_error}, \ref{fig:blog_all_error}).
% SGD-Shapley introduces two additional parameters: the initializations of the coefficients and the learning rate.

Finally, the last approach based on the least squares characterization is FastSHAP~\cite{jethani2021fastshap, covert2022learning}.
FastSHAP learns a separate
% deep
model (an \textit{explainer}) to estimate Shapley values in a single forward pass, and it is trained by amortizing the WLS problem (Eq.~\ref{eq:least_squares}) across many data examples. As a consequence of its WLS training objective, the globally optimal estimation model is a function that outputs exact Shapley values.
% FastSHAP amortizes the solution to the WLS problem by optimizing an explanation model with an objective for which the global optimizer would output exact Shapley values.  In this approach, the key assumption is that the explanation model well approximates the global optimizer. 
Since the explanation model will generally be non-optimal, the resultant estimates offer imperfect accuracy and are random across separate training runs
% in general imperfect
% biased and stochastic 
\cite{jethani2021fastshap}.  The major advantage of FastSHAP is that developers can frontload the cost of 
% sampling subsets and
training the explanation model, thereby providing subsequent users with fast Shapley value explanations.  

The fourth characterization of the Shapley value is based on a \textbf{multilinear extension} of the game~\cite{owen1972multilinear,okhrati2021multilinear}.  The multilinear extension extends a coalitional game to be a function on the $d$-cube $[0, 1]^d$ that is linear separately in each variable. Based on an integral of the multilinear extension's partial derivatives, the Shapley value can be defined as
% defines a function of a game that expands the domain of the game to the entirety of the D-cube $1^D$ and is linear in each variable.
% Based on this characterization, the Shapley value can be defined as:

\begin{equation}
\phi_i(v) = \int_0^1 g_i(q) dq,
\end{equation}

\noindent where $g_i(q)=\mathbb{E}[v(G_i \cup \{i\})-v(G_i)]$ and $G_i$ is a random subset of $D\setminus\{i\}$, with each feature having probability $q$ of being included.
% in the subset.
Perhaps surprisingly, as with the random order value characterization, estimating this formulation involves averaging many marginal contributions where the subsets are effectively drawn from $P(S) = \frac{|S|!(|D|-|S|-1)!}{|D|!}$.
% As such, 
% approaches based on
% both of these approaches can be viewed as techniques to draw subsets for ApproSemivalue,
% but we categorize them here based on the characterization that directly inspired each approach.

% This characterization is similar to the random order value characterization, but it can be viewed as an alternative way to draw from $P(S)$ in the semivalue characterization.

Based on this characterization, \textcite{okhrati2021multilinear} introduced an unbiased sampling-based estimator that we refer to as multilinear extension sampling.  The estimation consists of (1)~sampling a $q$ from the range $[0,1]$, and then (2)~sampling random subsets based on $q$ and evaluating the marginal contributions.   This procedure introduces an additional parameter, which is the balance between the number of samples of $q$ and the number of subsets $E_i$ to generate
% estimate
for each value of $q$.  The original version
% paper
draws 2 random subsets for each $q$, where $q$ is sampled at fixed intervals according to the trapezoid rule~\cite{okhrati2021multilinear}.  Finally, in terms of variants, \textcite{okhrati2021multilinear} find that antithetic sampling improves convergence, where for each subset they also compute the marginal contribution for the inverse subset.

To summarize, there are three main characterizations of the Shapley value from which unbiased, stochastic estimators have been derived: random order values, least squares values, and multilinear extensions.  Within each approach, there are a number of variants.  (1)~\textit{Adaptive sampling}, which has only been applied to IME (per-feature random order), but can easily be applied to a version of multilinear extension sampling that explains features independently.
% \footnote{It is possible to apply adaptive sampling to ApproShapley, but it would be much more complicated because it would only be beneficial when dropping many contiguous sets of features.}
(2)~\textit{Efficient sampling}, which aims to carefully draw samples 
% in a non-independent manner
% often anti-correlated subsets
to improve convergence over independent
% random
sampling.  In particular, one version of efficient sampling, antithetic sampling, 
% can be implemented within multiple approaches
is easy to implement and effective;
it has been applied to ApproShapley, KernelSHAP, and multilinear extension sampling, and it can also be easily extended to IME\footnote{The approach of drawing an inverse subset for each independently drawn subset was introduced as ``paired sampling'' in KernelSHAP~\cite{covert2020improving}, ``antithetic sampling'' for ApproShapley~\cite{mitchell2021sampling}, and ``halved sampling'' for multilinear extension sampling~\cite{okhrati2021multilinear}.  We refer to all of these approaches as ``antithetic sampling.''~\cite{rubinstein2016simulation}}.  Although the other efficient sampling techniques have mainly been examined in the context of ApproShapley, similar benefits may exist for IME, KernelSHAP, and multilinear extension sampling.  Finally, there is (3)~\textit{amortized explanation models}, which have only been applied to the least squares characterization~\cite{jethani2021fastshap}, but may be extended to other characterizations that where the Shapley value can be viewed
% is interpreted
as the solution to an optimization problem.
% the random order and multilinear extension characterizations as well.

% In terms of variants of each approach, we can roughly categorize them into the following categories: (1)~adaptive sampling, (2)~efficient sampling, and (3)~amortized explanation models.  In particular, adaptive sampling has only been applied to the random-order approximation of the Shapley values~\cite{vstrumbelj2014explaining}, but it can easily be extended to the multilinear extension approximation~\cite{okhrati2021multilinear} because they both explain one feature at a time.  Efficient sampling aims to draw an often correlated group of coalition samples for which the estimates of the Shapley value explanations have lower variance compared to random Monte Carlo sampling.  In particular, one version of efficient sampling, antithetic sampling, is easy to implement and effective; it has been applied to ApproShapley, multilinear extension sampling, and KernelSHAP, but can also be easily extended to IME.  Although the other efficient sampling techniques have mainly been examined in the context of ApproShapley, similar benefits may exist for IME, multilinear extension sampling, and KernelSHAP.  Finally, the amortized explanation models have only been developed for the least squares characterization of the Shapley value~\cite{jethani2021fastshap}, but may be extended to random order and multilinear extension characterizations as well.

\subsubsection{Empirically comparing model-agnostic approaches}

\begin{figure}[!ht]
\includegraphics[width=\textwidth]{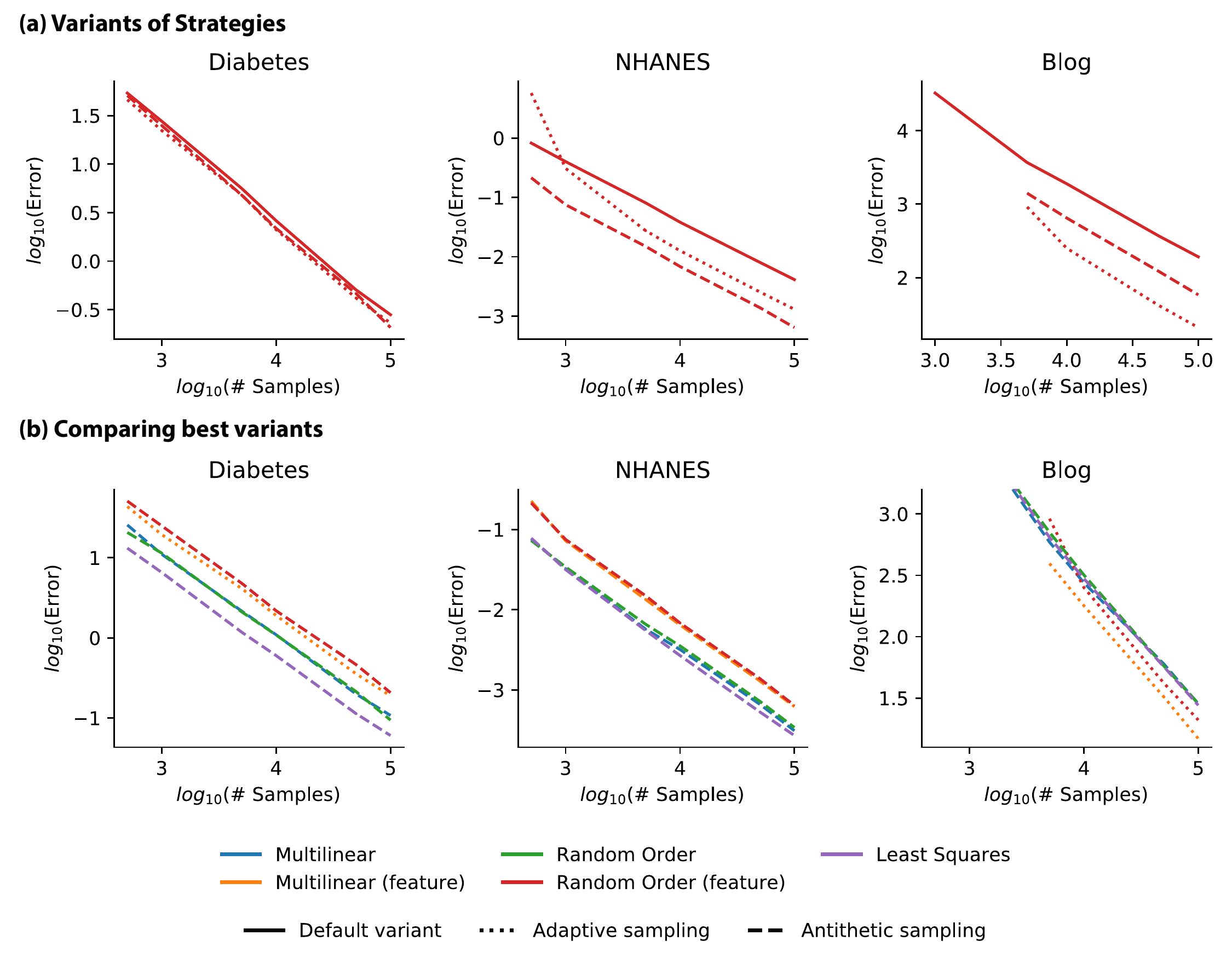}
\centering
\caption{Benchmarking unbiased, model-agnostic algorithms to estimate baseline Shapley values for a single explicand and baseline on XGB models with 100 trees.  
% Note that these methods are model-agnostic, so they can also be used to estimate marginal, conditional, and causal Shapley values.  
For simplicity, we calculate baseline Shapley values for all methods
% we focus on baseline Shapley values
because we aim to evaluate the tractable estimation strategy rather than the feature removal approach.
% In particular, the stochastic estimators include each sampling-based approach, where the prefixes MEF (multilinear extension, feature-wise), ROF (random order, feature-wise), RO (random order), and LS (least squares) correspond to multilinear extension sampling~\cite{okhrati2021multilinear}, IME~\cite{strumbelj2010efficient}, ApproShapley~\cite{castro2009polynomial}, and KernelSHAP~\cite{lundberg2017unified} respectively.  The suffixes ``ANTI'' and ``ADAPT'' stand for antithetic sampling and adaptive sampling respectively.  
In particular, the stochastic estimators include each sampling-based approach, where multilinear, random order, random order (feature), and least squares correspond to multilinear extension sampling~\cite{okhrati2021multilinear}, ApproShapley~\cite{castro2009polynomial}, IME~\cite{strumbelj2010efficient}, and KernelSHAP~\cite{lundberg2017unified} respectively.  Multilinear (feature) is a new approach based on the multilinear extension sampling approach which explains one feature at a time.  In addition, some methods are variants: either antithetic
% sampling
or adaptive sampling.  
On the x-axis we report the number of samples (subsets) used for each estimate, and on the y-axis we report the MSE relative to the true baseline Shapley value for 100 estimates with that many samples.  We use three real-world datasets: diabetes (10 features, regression), NHANES (79 features, classification), blog (280 features, regression).  For some variants, no error is shown for small numbers of samples; this is because each approach requires a different minimum number of samples to produce estimates for each feature.  (a) Variants of the random order, feature-wise strategy.  We report the variants for all four strategies in Appendix Figures \ref{fig:diabetes_all_error}, \ref{fig:nhanes_all_error}, \ref{fig:blog_all_error}.  (b) Benchmarking the most competitive variant of each stochastic estimator chosen according to the lowest error for $10^5$ samples.  Note that the full blog error plot (Appendix Figure~\ref{fig:blog_top_error}) is truncated to better showcase differences.  Finally, the MSE can be decomposed into bias and variance terms~\cite{covert2020improving}, which we show in Appendix Tables \ref{tab:diabetes_bias}, \ref{tab:nhanes_bias}, \ref{tab:blog_bias}, \ref{tab:diabetes_var}, \ref{tab:nhanes_var}, \ref{tab:blog_var}.}
\label{fig:shapley_algorithms}
\end{figure}

In Figure~\ref{fig:shapley_algorithms}, we examine the convergence of the four main
% unbiased
stochastic estimators (IME, ApproShapley, KernelSHAP and multilinear extension sampling) as well as several popular variants of these approaches (adaptive and antithetic sampling\footnote{The other efficient sampling techniques and amortized explanation models are more complex and out of the scope for this review.}).  We experiment with three datasets with varying numbers of features, all using XGBoost models.  To calculate convergence, we calculate estimation error (mean squared error) of estimates of baselines Shapley values from the true baseline Shapley values computed using Interventional TreeSHAP.  We also introduce four new approaches: IME with antithetic sampling, a new version of multilinear extension sampling that explain one feature at a time\footnote{The initial version of multilinear extension sampling~\cite{okhrati2021multilinear} explains all features simultaneously.  This enables re-use of model evaluations, but prohibits the use of adaptive sampling.}, and antithetic/adaptive sampling variants of this new multilinear extension sampling approach.

For the diabetes and NHANES datasets, which have
% only
10 and 79 features respectively, the anthithetic version of KernelSHAP
% offers the fastest convergence
converges fastest to the true Shapley values.
However, for the blog dataset, which has 280 features, we observe that IME and multilinear (feature) converge fastest, likely due to their use of adaptive sampling.  
% The multilinear extension sampling may be more effective because we draw samples of $q$ according to the trapezoid rule. 
Based on this finding, we hypothesize that adaptive sampling is important
% effective
in the presence of many features, because with many features there are more likely to be features without interaction effects or with near-zero contributions.  Such features will have little to no variance in their marginal contributions.  We tested this
% test our
hypothesis by modifying the diabetes dataset by adding 100 zero features that
% will
have no impact on the model and therefore no interaction effects.  In this scenario, we find that the dummy features slow convergence and that the adaptive sampling approaches are least affected, likely because they can determine which features
% quickly
converge rapidly and then ignore them (Appendix Figures \ref{fig:diabetes_zeros_all_error}, \ref{fig:diabetes_zeros_top_error}).

Finally, an important
% additional
desideratum of feature attribution estimates is the \textit{efficiency axiom}~\cite{shapley1953value}, which means that the attributions sum
% add up
to the game's value
% value of the game
with all players minus the value with none (Figure~\ref{fig:shapley_values}b).  In terms of the sampling-based
% unbiased
estimators, only ApproShapley and KernelSHAP are guaranteed to satisfy the efficiency property.  However, it is possible to adjust attributions by evenly splitting the efficiency gap between all features using the \textit{additive efficient normalization} operation~\cite{ruiz1998family}.  This
% type of
normalization step is used to ensure that the efficiency property holds
% efficiency during optimization
for both FastSHAP and
% a version of
SGD-Shapley~\cite{jethani2021fastshap,simon2020projected}.  It can also be used to ensure efficiency as a final post-processing step for IME and multilinear extension sampling, and it is guaranteed to improve estimates in terms of Euclidean distance to the true Shapley values
% attributions
without affecting the estimators' bias~\cite{jethani2021fastshap}.

% Alex: Published examples of default parameters, that are being ignored and may not be appropriate.

% \hugh{Include comparison of the above three approaches}

These model-agnostic strategies for estimating Shapley values are appealing because they are flexible: they can be applied to any coalitional game and therefore
% subsequently
any machine learning model.  However, one major downside of these
% approximation-based
approaches is that they are inherently stochastic.  Although most methods are guaranteed to be correct given an infinite number of samples (i.e., they are consistent estimators),
% deterministic given an exponential number of samples,
users have finite
% much smaller
computational budgets, leading to estimators with potentially non-trivial variance.  In response, some
% papers
methods utilize techniques to forecast and detect convergence when the estimated variance drops below a fixed threshold~\cite{covert2020understanding, covert2020improving}.  However, even with convergence detection, model-agnostic
% sampling-based
% approximation-based
explanations can be prohibitively expensive.
% (with the exception of FastSHAP, which produces efficient explanations after training)
Motivated in part by the computational complexity of these methods, a number of approaches have been developed to estimate Shapley value explanations more efficiently by making
% based on
assumptions about the type of model being explained.

\subsubsection{Model-specific approaches}
\label{sec:model_specific_approaches}

In terms of model-specific approaches, algorithms have been designed for several
% each of the following
popular model types: linear models, tree models, and deep models.  These approaches are less flexible than model-agnostic approaches, in that they assume a specific model type and often
% in many cases
a specific feature removal approach, but they are generally significantly
% drastically
faster to calculate.  

First, one of the simplest model types to explain is \textbf{linear models} (also discussed in Section~\ref{sec:shapley_value_feature_attributions}). 
% which often serve as a sanity check for which many explanation methods coincide~\cite{ancona2019explaining}.
For linear models, baseline
% Shapley values
and marginal Shapley values have a closed-form solution based on the coefficients of the linear model ($\beta$) and the values of the explicand ($x^e$) and the baseline(s).
% (LinearSHAP~\cite{lundberg2017unified,vstrumbelj2014explaining}):
If we let $E$ represent a set of baseline values for marginal Shapley values, or $E = \{x^b\}$
for baseline Shapley values,
% for a single fixed baseline,
we can write the mean value for feature $x_i$ as $\mu_i = \frac{1}{|E|} \sum_{x^b \in E} x^b$.  Then, LinearSHAP~\cite{lundberg2017unified,vstrumbelj2014explaining} gives the following result for the marginal or baseline Shapley values:

\begin{equation}
    \phi_i(f,x^e) = \beta_i(x^e_i - \mu^b_i).
\end{equation}

% where $\mu^b_i$ is the average value of feature $x_i$ across all the baselines.
Computing these values is straightforward, and it
% exact estimates
has linear complexity in the number of features, versus exponential complexity in the general case. 
% linear complexity in the model size and the number of baselines. 
Interestingly, for linear models the marginal Shapley value is equivalent to the baseline Shapley value with a mean baseline, which is why the mean baseline may be a good choice for linear models (see Section \ref{sec:shapley_linear_examples}).

Alternatively, correlated LinearSHAP~\cite{chen2020true} estimates conditional Shapley values for linear models assuming that the data follows a \textit{multivariate Gaussian distribution}.
% given multivariate normal data.
The key idea behind this approach is that for linear models,
% Conditional Shapley values are easy to estimate in this setting because for linear models,
% a linear model $f$,
the expectation of the model's prediction equals the model's prediction for the average sample, or $\mathbb{E}[f(x) \mid x_S] = f(\mathbb{E}[x \mid x_S])$), which is not true in general.
% Then, there is a closed form estimate for the conditional expectation $\mathbb{E}[x \mid x_S]$ if we assume normally distributed data.  The final estimates are:
Using the closed-form solution for $\mathbb{E}[x \mid x_S]$ given multivariate normal data, we can calculate the conditional Shapley values as

\begin{equation}
    \phi_i(f,x^e) = \beta^\top A_i \mu + \beta^\top B_i x^e,
\end{equation}

\noindent where $\mu$ is the mean feature vector
% of the baselines,
and $A_i$ and $B_i$ are summations over an exponential number of coalitions (see~\cite{chen2020true} for more details).  In practice,
% estimates for
$A_i$ and $B_i$ are estimated using
% formed by
a sampling-based procedure.
% a subsample of all coalitions;
Then, subsequent explanations are extremely fast to compute regardless of the explicand value.  This approach resembles
% Note the similarity to
FastSHAP~\cite{jethani2021fastshap} because both approaches incur most of their
% the bulk of the
computation up-front and then provide very fast explanations.  Correlated LinearSHAP is biased if the data does not follow a multivariate Gaussian distribution,
% assumption of normality does not hold
which is rarely the case
% it rarely will
% it often will not
in practice, and it is stochastic because $A_i$ and $B_i$ are estimated.

Next, another popular model type is \textbf{tree models}.  Tree models include
% models such as
decision trees, as well as ensembles like random forests and gradient boosted trees.  These
% models
are more complex than linear models and they can represent
% comprise
non-linear and interaction effects, but perhaps surprisingly, it is possible to
% exactly
calculate baseline and marginal Shapley values exactly (Interventional TreeSHAP) and approximate conditional Shapley values (Path-dependent TreeSHAP) tractably for such models.
% Perhaps surprisingly, it is 
% % Despite their complexity, it is surprisingly
% possible to exactly compute baseline
% % Shapley values
% and marginal Shapley values for tree-based models. 

Interventional TreeSHAP is an algorithm that exactly calculates baseline and marginal Shapley values in time linear in the size of the tree model and the number of baselines~\cite{lundberg2020local}.
This is possible because a tree model can be 
% equivalently
represented as a disjoint set of outputs for each leaf in the tree.  Then, each leaf's contribution to any given feature's Shapley value can be computed at the leaves of the tree assuming a coalitional game whose players are the features along the path from the root to the current leaf.
Using a dynamic programming algorithm,
% By using dynamic programming,
Interventional TreeSHAP
computes the Shapley value explanations for all features simultaneously by iterating through the nodes in the tree.

Then, Path-dependent TreeSHAP is an algorithm designed to estimate conditional Shapley values, where the conditional expectation is approximated by the structure of the tree model~\cite{lundberg2020local}.
% Based on the set of present and absent features,
Given a set of present features,
the algorithm handles internal nodes for absent features by traversing each branch in proportion to how many examples in the dataset follow each direction.
% either traverses a branch in the tree according to the explicand or averages over both branches in order to estimate the conditional expectation.
This algorithm can be viewed as an application of Shapley cohort refinement~\cite{mase2019explaining}, where the cohort is defined by the preceding nodes in the tree model and the baselines are the entire training set.  In the end, it is possible to estimate a biased, variance-free version of conditional Shapley values in $O(LH^2)$ time, where $L$ is the number of leaves and $H$ is the depth of the tree.  Path-dependent TreeSHAP is a biased estimator for conditional Shapley values because its estimate of the conditional expectation is
% will be
imperfect.

In comparison to Interventional TreeSHAP, Path-dependent TreeSHAP does not have a linear dependency on the number of baselines, because it utilizes node weights that represent the portion of baselines that fall on each node (based on the splits in the tree).  Finally, in order to incorporate a tree ensemble, both approaches calculate
% estimate
explanations separately for each tree in the ensemble and then combine them linearly.  This yields exact estimates for baseline and marginal Shapley values, because the Shapley value is additive with respect to the model~\cite{lundberg2020local}.

Finally, another popular but opaque class of models are \textbf{deep models} (i.e., deep neural networks).  Unlike for linear and tree models, we are unaware of any
% deep model specific
approach to estimate conditional Shapley values for deep models, but we discuss several
% three
approaches that
% designed to
estimate baseline and marginal Shapley values.

One early method to explain deep models, called DeepLIFT, was designed to propagate
% developed to explain deep models by propagating 
attributions through a deep network for a single explicand and baseline~\cite{shrikumar2017learning}.  DeepLIFT propagates activation differences through each layer in the deep network, while maintaining the Shapley value's efficiency property at each layer using a chain rule based on either a Rescale rule or a RevealCancel rule, which can be viewed as approximations of the Shapley value~\cite{chen2021explaining}.  Due to the chain rule and these local approximations, DeepLIFT produces biased estimates of baseline Shapley values.  Later, an extension of this method named DeepSHAP was designed to produce biased estimates of marginal Shapley values~\cite{chen2021explaining}.  Despite its bias, DeepSHAP is useful because the computational complexity is on the order of the size of the model and the number of baselines, and the explanations have been shown to be useful empirically~\cite{reiter2020developing,chen2021explaining}.  In addition, the Rescale rule is general enough to propagate attributions through pipelines of linear, tree, and deep models~\cite{chen2021explaining}.

Another method to estimate baseline Shapley values for deep models is Deep Approximate Shapley Propagation (DASP)~\cite{ancona2019explaining}.  DASP utilizes uncertainty propagation to estimate baseline Shapley values.
% In order
To do so, the authors rely on a definition of the Shapley value that averages the expected marginal contribution for each coalition size.
% each coalition size's expected marginal contribution.
For each coalition size $k$ and a zero baseline, the input distribution from the random coalitions is modeled as a normal random variable whose parameters are a function of $k$.  Since the input distributions are normal random variables, it is possible to propagate uncertainty for specific layers by matching first and second-order central moments and thereby estimate each expected marginal contribution.  
% The moments can be matched easily for affine transformations, ReLU activations, and max pooling because closed-form solutions are known.  
Based on an empirical study,
% empirical findings,
DASP produces baseline Shapley values estimates with lower bias than
% in comparison to
DeepLIFT~\cite{ancona2019explaining}.  However, DASP is more computationally costly
% expensive computationally
and requires up to $O(d^2)$ model evaluations, where $d$ is the number of features.  
% In practice, DASP converges with fewer evaluations, but still requires a minimum of $O(d)$ model evaluations.  
Although DASP is deterministic (variance-free) with $O(d^2)$ model evaluations, it is biased because the moment propagation relies on an assumption of independent inputs that is violated at internal nodes whose inputs are given by the previous layer's outputs.
% the outputs of the previous layer.  
% In addition, DASP assumes the first and second moments can be matched for all layers in the network, but it is unclear whether closed form solutions exist for more complicated layers.  In particular, DASP describes moment matching for affine transformations, ReLU activations, and max pooling layers.

One final method to estimate baseline Shapley values for deep models is Shapley Explanation Networks (ShapNets)~\cite{wang2020shapley}.  ShapNets restrict the deep model to have a
% be of a
specific architecture for which baseline Shapley values are easier to estimate.  The authors make a stronger assumption than DASP or DeepLIFT/DeepSHAP by not only restricting the model to be a neural network, but by requiring a specific architecture
% to have a very specific architecture
where hidden nodes have a small input dimension $h$ (typically between 2-4).  In this setting, ShapNets can construct baseline Shapley values for each hidden node because the exponential cost is low for small $h$.  The authors present two methods that follow the architecture assumption.
% comply to this restriction.
(1)~\textit{Shallow ShapNets}: networks that have a single hidden layer, and where baseline Shapley values can be calculated exactly.
% estimates for the entire network can be exactly computed.
Although they are easy to explain, these networks suffer in terms of model capacity and have lower predictive accuracy than
% compared to
other deep models.  (2)~\textit{Deep ShapNets}: networks with multiple layers through which we can
% they
calculate explanations hierarchically.  For Deep ShapNets, the final estimates are
% will be
biased because of this hierarchical, layer-wise procedure.
% cascading.
However, since Deep ShapNets can have multiple layers, they are more performant in terms of making predictions, although they are still
% may still be
more limited than standard deep models.
% in terms of capturing feature interactions.
An additional advantage of ShapNets is that they enable developers to regularize explanations
% explanation regularization so developers can penalize explanations
% regularize
based on prior information without a costly estimation procedure~\cite{wang2020shapley}.  
% In contrast, other approaches to estimate Shapley value explanations cannot easily be regularized because they are generally not differentiable.

DASP and Deep ShapNets are originally designed to estimate baseline Shapley values with a zero baseline: DASP assumes a zero baseline to obtain an appropriate input distribution, and Deep ShapNets uses zero baselines in internal nodes.  However, it may be possible to adapt DASP and Deep ShapNets to use arbitrary baselines (as in DeepLIFT and Shallow ShapNets), in which case it would be possible to estimate marginal Shapley values as DeepSHAP does.
In terms of computational complexity, DeepLIFT, Shallow ShapNets, and Deep ShapNets can estimate baseline Shapley values with a constant number of model evaluations (for a fixed $h$).  In contrast, DASP requires a minimum of $d$ model evaluations and up to $O(d^2)$ model evaluations for a single estimate of baseline Shapley values.

A final difference between these approaches is in their assumptions.  Shallow ShapNets and Deep ShapNets make the strongest assumptions by restricting the deep model's architecture.
% of the deep model.
DASP makes a strong assumption that we can
% it is possible to
perform first and second-order central moment matching for each layer in the deep model, and the original work only describes moment matching for affine transformations, ReLU activations and max pooling layers.  Finally, DeepLIFT and DeepSHAP assume deep models, but they are flexible and support more types of layers than DASP or ShapNets.  However, as a consequence
% repercussion
of DeepLIFT's flexibility, its baseline Shapley value estimates
% empirically
have higher bias compared
% in comparison
to DASP or ShapNets~\cite{ancona2019explaining,wang2020shapley}.

% \subsection{(Aside) Incorporating prior knowledge with causal structure}

% \begin{itemize}
%     \item Incorporating causal structure can often serve to increase credit given to variables that are known to be causal parents or ancestors of other nodes.  Three prominent examples include:
%     \begin{itemize}
%         \item Asymmetric Shapley values which modify the weighting scheme of Shapley values to accomodate a bipartite set of ancestors and descendents
%         \item Shapley flow which attributes credit to edges in a causal graph rather than features
%         \item Causal Shapley which uses the marginal expectation with more complicated causal structures
%     \end{itemize}
%     \item \textbf{These methods require an additional causal graph parameter which is generally very difficult to estimate from data (and even if it is fully specified the marginal expectation may not be identifiable).  As such, they primarily utilize causal graphs derived from prior knowledge.}
%     \item In addition, these causal Shapley values require estimating a potentially exponential number of conditional expectations and thus introduce the causal graph parameter in addition to the distribution parameter.
% \end{itemize}

\section{Discussion}

% \hugh{Rewrite to add in recommendations: linear models, GAMs don't really need marginal Shapley values which are kind of built in.  Tree models, use marginal Shapley values for sure.  Deep models, three imperfect approaches to estimate marginal Shapley values.  For linear and tree there are approaches that approximate conditional Shapley values, but they are not perfect.}

In this work, we provided a detailed overview of numerous algorithms for generating Shapley value explanations. In particular, we delved into the two main factors of complexity underlying
% Shapley value
such explanations: the feature removal approach and the tractable estimation strategy.
% Understanding the two factors of complexity underlying Shapley value explanations, the feature removal approach and the estimation strategy, allows us to more easily understand existing explanation techniques.  
Disentangling the complexity in the literature into these two factors allows us to more easily understand the key innovations
% and differences
in recently proposed approaches.

In terms of feature removal approaches, algorithms that aim to estimate baseline Shapley values are generally unbiased, but choosing a single baseline to represent feature removal is challenging.  Similarly, algorithms that aim to estimate marginal Shapley values will also generally be unbiased in their Shapley value estimates.  Finally, algorithms that aim to estimate conditional Shapley values will be biased because the conditional expectation is fundamentally challenging
% hard
to estimate.  Conditional Shapley values are currently difficult to estimate with low bias and variance, except in the case of linear models; however, depending on the use case, it may be preferable to use an imperfect approximation rather than switch to baseline or marginal Shapley values.

In terms of the exponential complexity of Shapley values, model-agnostic approaches are often more flexible and bias-free, but they produce estimators with non-trivial variance.  By contrast, model-specific approaches are typically deterministic and sometimes unbiased.  Of the model-specific methods, only LinearSHAP and Interventional TreeSHAP have no bias for baseline and marginal Shapley values.  In particular, we find that the Interventional TreeSHAP explanations are fairly remarkable for being non-trivial, bias-free, and variance-free.  As such, tree models including decision trees, random forests, and gradient boosted trees are particularly well-suited to Shapley value explanations.

Furthermore, based on the feature removal approach and estimation strategy of each approach, we can understand the sources of bias and variance within many existing algorithms (Table \ref{tab:methods}).  IME~\cite{strumbelj2010efficient}, for instance, is bias-free, because marginal Shapley values and the random order value estimation strategy are both bias-free.  However, IME estimates have non-zero variance
% are variable
because the estimation strategy is stochastic
% a stochastic approximation
(random order value sampling).  In contrast, Shapley cohort refinement estimates~\cite{mase2019explaining} have both non-zero bias and
% are both biased and have
non-zero variance.
% variable.
Their bias comes from modeling the conditional expectation using an empirical, similarity-based approach, and the variance comes from the sampling-based estimation strategy (random order value sampling).

In practice, Shapley value explanations are widely used in both industry and academia.  Although they are powerful tools for explaining models, it is important for users to be aware of important parameters associated with the algorithms used to estimate them.  In particular, we
% would
recommend that any analysis based on Shapley values should report parameters including the type of Shapley value explanation (the feature removal approach), the baseline distribution used to estimate the coalitional game, and the estimation strategy.  For sampling-based strategies, it is important for users to include a discussion of convergence in order to validate their feature attribution estimates.  Finally, developers of Shapley value explanation tools should strive to be transparent about convergence by explicitly performing automatic convergence detection.  Convergence results based on the central limit theorem are straightforward for the majority of model-agnostic estimators we discussed, although they are not always implemented in public packages.  Note that convergence analysis
% and detection
is more difficult
% harder
for the least squares estimators, but
% where the variance is hard to estimate.
\textcite{covert2020improving} discuss this issue and present a convergence detection approach for KernelSHAP.

Future research directions include investigating new stopping conditions for convergence detection.  Existing work proposes stopping once the largest standard deviation is smaller than a prescribed threshold~\cite{covert2020improving}, but depending on the threshold, the variance may still be high enough that the relative importance of features can change.  Therefore, a new stopping condition could be when additional marginal contributions are highly unlikely to change the relative ordering of attributions for all features.   Another important future research direction is Shapley value estimation for deep models.  Current model-specific approaches to explain deep models are biased, even for marginal Shapley values, and no model-specific algorithms exist to estimate conditional Shapley values.  One promising model-agnostic approach is FastSHAP~\cite{jethani2021fastshap,covert2022learning}, which speeds up explanations using an explainer model, although it requires
% incurs
a large upfront cost to train this model.  Finally, because approximating the conditional expectation for conditional Shapley values is so hard, it constitutes an important future research direction that would benefit from new methods or systematic evaluations of existing approaches.

% \begin{tcolorbox}[before upper={\parindent15pt}]
% \noindent \textbf{Recommendations based on data domain}
\section{Recommendations based on data domain}

What we have discussed in this paper is largely agnostic to the type of data.  However, there are a few characteristics of the data being analyzed that may change the best practices for generating
% to compute
Shapley value explanations.

The first characteristic is the number of features.  Models with more input features will be more computationally expensive to explain.  In particular, for model-agnostic algorithms, as the number of features increases, the total number of possible coalitions increases exponentially.  In this setting it may be valuable to reduce the number of features by carefully filtering the ones which do not vary or are highly redundant with other features.  This type of feature selection can even be performed prior to model-fitting and is already a common practice.

The second characteristic is the number of samples.  The number of samples likely plays a larger role in fitting the original predictive model than in generating
% estimating
the explanation.  However, for conditional Shapley values, having a large number of samples is important for
% to
creating accurate estimates of the conditional expectations/distributions.  If the number of samples is very low, it may be better to rely on parametric assumptions or use marginal Shapley values instead.

The third characteristic is the feature correlation.  In highly correlated settings, one can expect larger discrepancies between marginal and conditional Shapley values.  Then, carefully choosing the feature removal approach or comparing estimates of both marginal and conditional Shapley values can be valuable.  Highly correlated features can also make it harder to understand feature attributions in general.  For images in particular, the importance of a single pixel may not be semantically meaningful in isolation.  In these cases, it may be useful to use explanation methods that aim to understand higher level concepts~\cite{koh2020concept}.

Beyond correlated features, there may also be structure within the data.  Tabular data are typically considered unstructured whereas image and text data is structured because neighboring pixels and words are strongly correlated.  For tabular data, it may be best to use tree ensembles (gradient boosted trees or random forests) which are both performant and easy to explain using the two versions fo TreeSHAP (Interventional and Path-dependent)~\cite{lundberg2020local}.  For structured data, it may be valuable to use methods such as L-Shapley and C-Shapley that are designed to estimate Shapley value explanations more tractably in structured settings~\cite{chen2018shapley}.  Furthermore, grouping features may be natural for structured data (e.g., superpixels for image data or sentences/n-grams for text data), because it greatly reduces the computational complexity of most algorithms.  Finally, since structured data often calls for complex deep models, which are expensive to evaluate, methodologies such as FastSHAP can be useful for accelerating explanations.

The fourth characteristic is prior knowledge of causal relationships.  Although causal knowledge is unavailable
% unknown
for the vast majority of datasets, it can be used to generate Shapley value explanations that better respect causal relationships~\cite{heskes2020causal,frye2020asymmetric} or generate explanations that assign importance to edges in the causal graph~\cite{wang2021shapley}.  These techniques may be a better alternative to conditional Shapley values, which respect the data manifold, because they respect the causal relationships underlying the correlated features.

The fifth characteristic is whether there is a natural interpretation of absent features.  For certain types of data, there may be preconceived notions of feature absence.  For instance, in text data it may be natural to remove features from models that take variable length inputs or use masking tokens to denote feature removal.  In images, it is often common to assume some form of gray, black, or blurred baseline; these approaches are somewhat dissatisfying because they are data-specific notions of feature removal.  However, given that model evaluations are often exorbitantly expensive in these domains, these techniques may provide simple, yet tractable alternatives to marginal or conditional Shapley values\footnote{Note that for the conditional expectation, surrogate models are tractable once they are trained~\cite{frye2020shapley,jethani2021fastshap} because they directly estimate the conditional expectation in a single model evaluation.  However, using a surrogate requires training or fine-tuning an additional model.
% the surrogate requires an initial training cost that is non-trivial.
% initial cost of training the surrogate model is expensive.
}.

\section{Related work}

In this paper, we focused on describing popular algorithms to estimate local feature attributions based on the Shapley value.  However, there are a number of adjacent explanation approaches that are not the focus of this discussion.  Two broad categories of such approaches include alternative definitions of coalitional games, and different game-theoretic solution concepts.

We focus on three popular coalitional games where the players represent
% are analogous to
features and the value is the model's prediction for a single example.  However, as discussed by \textcite{covert2021explaining}, there are several methods that use different
% a variety of alternative values used by
coalitional games, including global feature attributions where the value is the model's mean test loss~\cite{covert2020understanding}, and
% or even
local feature attributions where the value is the model's per-sample loss~\cite{lundberg2020local}. 
Other examples include games where the value is the maximum flow of attention weights in transformer models~\cite{ethayarajh2021attention},
% games
where the players are analogous to samples in the training data~\cite{ghorbani2019data},
% games
where the players are analogous to neurons in a deep model~\cite{ghorbani2020neuron}, and
% games
where the players are analogous to edges in a causal graph~\cite{wang2021shapley}.  Although these methods are largely outside the scope of this paper, a variety of applications of the Shapley value in machine learning are discussed in \textcite{rozemberczki2022shapley}.

Secondly, there are game-theoretic solution concepts beyond the Shapley value that can be utilized to explain machine learning models.  The first method, named asymmetric Shapley values, is designed to generate feature attributions that incorporate causal information~\cite{frye2020asymmetric}.  To do so, asymmetric Shapley values are based on random order values where weights are set to zero if they are inconsistent with the underlying causal graph.
% causal ordering.
Next, L-Shapley and C-Shapley (also discussed in Section~\ref{sec:model_agnostic_approaches}) are
% more
computationally efficient estimators for Shapley value explanations designed for structured data; they are technically \textit{probabilistic values}, a generalization of semivalues~\cite{chen2018shapley,monderer2002variations}.  Similarly, Banzhaf values, an alternative to Shapley values, are also semivalues, but each coalition is given equal weight in the summation (see Eq.~\ref{eq:shapley_definition}).  Banzhaf values have been used to explain machine learning models in a variety of settings~\cite{karczmarz2021improved,chen2020ls}.  Another solution concept designed to incorporate structural information about coalitions is the Owen value.  The Owen value has been used to design a hierarchical explanation technique named PartitionExplainer within the SHAP package\footnote{\url{https://shap.readthedocs.io/en/latest/generated/shap.explainers.Partition.html}} and as a way to accommodate groups of strongly correlated features~\cite{miroshnikov2021mutual}.  Finally, Aumann-Shapley values are an extension of Shapley values to infinite games~\cite{aumann2015values}, and they 
% Aumann-Shapley values
are connected to an explanation method named Integrated Gradients~\cite{sundararajan2017axiomatic}.  Integrated Gradients requires gradients of model's prediction with respect to the features, so it cannot be used for certain types of non-differentiable models (e.g., tree models, nearest neighbor models), and it represents features' absence in a continuous rather than discrete manner that typically requires a fixed baseline (similar to baseline Shapley values).

% Ian: this looks like a continuation of the paper, going to try to make it look separate

\clearpage

\section*{Data and code availability}

The diabetes dataset is publicly available (\url{https://www4.stat.ncsu.edu/~boos/var.select/diabetes.html}), and we use the version from the \texttt{sklearn} package.  The NHANES dataset is publicly available (\url{https://wwwn.cdc.gov/nchs/nhanes/nhefs/}), and we use the version from the \texttt{shap} package.  The blog dataset is publicly available (\url{https://archive.ics.uci.edu/ml/datasets/BlogFeedback}).

\section*{Code availability}

The code for the experiments is available here: \url{https://github.com/suinleelab/shapley_algorithms}.

\section*{Acknowledgements}

We thank Pascal Sturmfels, Joseph Janizek, Gabriel Erion, and Alex DeGrave for helpful discussions.  This work was funded by National Science Foundation [DBI-1759487, DBI-1552309, DGE-1762114, and DGE-1256082]; National Institutes of Health [R35 GM 128638, and R01 NIA AG 061132].

% \newpage

% \bibliography{main}
% \bibliographystyle{plainnat}

\vskip 0.2in
\printbibliography

@article{doshi2017towards,
  title={Towards a rigorous science of interpretable machine learning},
  author={Doshi-Velez, Finale and Kim, Been},
  journal={arXiv preprint arXiv:1702.08608},
  year={2017}
}

@article{lundberg2020local,
  title={From local explanations to global understanding with explainable AI for trees},
  author={Lundberg, Scott M and Erion, Gabriel and Chen, Hugh and DeGrave, Alex and Prutkin, Jordan M and Nair, Bala and Katz, Ronit and Himmelfarb, Jonathan and Bansal, Nisha and Lee, Su-In},
  journal={Nature Machine Intelligence},
  volume={2},
  number={1},
  pages={56--67},
  year={2020}
}

@article{young1985monotonic,
  title={Monotonic solutions of cooperative games},
  author={Young, H Peyton},
  journal={International Journal of Game Theory},
  volume={14},
  number={2},
  pages={65--72},
  year={1985},
  publisher={Springer}
}

@inproceedings{lundberg2017unified,
  title={A unified approach to interpreting model predictions},
  author={Lundberg, Scott M and Lee, Su-In},
  booktitle={Advances in Neural Information Processing Systems},
  pages={4765--4774},
  year={2017}
}

@article{chen2020true,
  title={True to the Model or True to the Data?},
  author={Chen, Hugh and Janizek, Joseph D and Lundberg, Scott and Lee, Su-In},
  journal={arXiv preprint arXiv:2006.16234},
  year={2020}
}

@inproceedings{ribeiro2016should,
  title={" Why should I trust you?" Explaining the predictions of any classifier},
  author={Ribeiro, Marco Tulio and Singh, Sameer and Guestrin, Carlos},
  booktitle={Proceedings of the 22nd ACM SIGKDD International Conference on Knowledge Discovery and Data Mining},
  pages={1135--1144},
  year={2016}
}

@inproceedings{shrikumar2017learning,
  title={Learning important features through propagating activation differences},
  author={Shrikumar, Avanti and Greenside, Peyton and Kundaje, Anshul},
  booktitle={Proceedings of the 34th International Conference on Machine Learning-Volume 70},
  pages={3145--3153},
  year={2017},
  organization={JMLR. org}
}

@article{sturmfels2020visualizing,
  title={Visualizing the Impact of Feature Attribution Baselines},
  author={Sturmfels, Pascal and Lundberg, Scott and Lee, Su-In},
  journal={Distill},
  volume={5},
  number={1},
  pages={e22},
  year={2020}
}

@inproceedings{fong2017interpretable,
  title={Interpretable explanations of black boxes by meaningful perturbation},
  author={Fong, Ruth C and Vedaldi, Andrea},
  booktitle={Proceedings of the IEEE International Conference on Computer Vision},
  pages={3429--3437},
  year={2017}
}

@article{mase2019explaining,
  title={Explaining black box decisions by Shapley cohort refinement},
  author={Mase, Masayoshi and Owen, Art B and Seiler, Benjamin},
  journal={arXiv preprint arXiv:1911.00467},
  year={2019}
}

@inproceedings{merrick2020explanation,
  title={The Explanation Game: Explaining Machine Learning Models Using Shapley Values},
  author={Merrick, Luke and Taly, Ankur},
  booktitle={International Cross-Domain Conference for Machine Learning and Knowledge Extraction},
  pages={17--38},
  year={2020},
  organization={Springer}
}

@inproceedings{janzing2020feature,
  title={Feature relevance quantification in explainable AI: A causal problem},
  author={Janzing, Dominik and Minorics, Lenon and Blobaum, Patrick},
  booktitle={International Conference on Artificial Intelligence and Statistics},
  pages={2907--2916},
  year={2020},
  organization={PMLR}
}

@article{aas2019explaining,
  title={Explaining individual predictions when features are dependent: More accurate approximations to Shapley values},
  author={Aas, Kjersti and Jullum, Martin and Loland, Anders},
  journal={arXiv preprint arXiv:1903.10464},
  year={2019}
}

@article{frye2020shapley,
  title={Shapley-based explainability on the data manifold},
  author={Frye, Christopher and de Mijolla, Damien and Cowton, Laurence and Stanley, Megan and Feige, Ilya},
  journal={arXiv preprint arXiv:2006.01272},
  year={2020}
}

@inproceedings{covert2020improving,
  title={Improving KernelSHAP: Practical Shapley value estimation using linear regression},
  author={Covert, Ian and Lee, Su-In},
  booktitle={International Conference on Artificial Intelligence and Statistics},
  pages={3457--3465},
  year={2021},
  organization={PMLR}
}

@article{strumbelj2010efficient,
  title={An efficient explanation of individual classifications using game theory},
  author={Strumbelj, Erik and Kononenko, Igor},
  journal={The Journal of Machine Learning Research},
  volume={11},
  pages={1--18},
  year={2010},
  publisher={JMLR. org}
}

@inproceedings{ancona2019explaining,
  title={Explaining deep neural networks with a polynomial time algorithm for Shapley value approximation},
  author={Ancona, Marco and Oztireli, Cengiz and Gross, Markus},
  booktitle={International Conference on Machine Learning},
  pages={272--281},
  year={2019},
  organization={PMLR}
}

@article{silver2017mastering,
  title={Mastering the game of {Go} without human knowledge},
  author={Silver, David and Schrittwieser, Julian and Simonyan, Karen and Antonoglou, Ioannis and Huang, Aja and Guez, Arthur and Hubert, Thomas and Baker, Lucas and Lai, Matthew and Bolton, Adrian and others},
  journal={Nature},
  volume={550},
  number={7676},
  pages={354--359},
  year={2017},
  publisher={Nature Publishing Group}
}

@article{moravvcik2017deepstack,
  title={Deepstack: Expert-level artificial intelligence in heads-up no-limit poker},
  author={Moravcik, Matej and Schmid, Martin and Burch, Neil and Lisy, Viliam and Morrill, Dustin and Bard, Nolan and Davis, Trevor and Waugh, Kevin and Johanson, Michael and Bowling, Michael},
  journal={Science},
  volume={356},
  number={6337},
  pages={508--513},
  year={2017},
  publisher={American Association for the Advancement of Science}
}

@article{vinyals2019grandmaster,
  title={Grandmaster level in StarCraft II using multi-agent reinforcement learning},
  author={Vinyals, Oriol and Babuschkin, Igor and Czarnecki, Wojciech M and Mathieu, Michael and Dudzik, Andrew and Chung, Junyoung and Choi, David H and Powell, Richard and Ewalds, Timo and Georgiev, Petko and others},
  journal={Nature},
  volume={575},
  number={7782},
  pages={350--354},
  year={2019},
  publisher={Nature Publishing Group}
}

@article{jumper2021highly,
  title={Highly accurate protein structure prediction with AlphaFold},
  author={Jumper, John and Evans, Richard and Pritzel, Alexander and Green, Tim and Figurnov, Michael and Ronneberger, Olaf and Tunyasuvunakool, Kathryn and Bates, Russ and Zidek, Augustin and Potapenko, Anna and others},
  journal={Nature},
  volume={596},
  number={7873},
  pages={583--589},
  year={2021},
  publisher={Nature Publishing Group}
}

@article{jean2014using,
  title={On using very large target vocabulary for neural machine translation},
  author={Jean, Sebastien and Cho, Kyunghyun and Memisevic, Roland and Bengio, Yoshua},
  journal={arXiv preprint arXiv:1412.2007},
  year={2014}
}

@article{lecun2015deep,
  title={Deep learning},
  author={LeCun, Yann and Bengio, Yoshua and Hinton, Geoffrey},
  journal={Nature},
  volume={521},
  number={7553},
  pages={436--444},
  year={2015},
  publisher={Nature Publishing Group}
}

@inproceedings{chen2016xgboost,
  title={XGBoost: A scalable tree boosting system},
  author={Chen, Tianqi and Guestrin, Carlos},
  booktitle={Proceedings of the 22nd ACM SIGKDD International Conference on Knowledge Discovery and Data Mining},
  pages={785--794},
  year={2016}
}

@article{breiman2001random,
  title={Random forests},
  author={Breiman, Leo},
  journal={Machine Learning},
  volume={45},
  number={1},
  pages={5--32},
  year={2001},
  publisher={Springer}
}

@inproceedings{steinkraus2005using,
  title={Using GPUs for machine learning algorithms},
  author={Steinkraus, Dave and Buck, Ian and Simard, PY},
  booktitle={Eighth International Conference on Document Analysis and Recognition (ICDAR'05)},
  pages={1115--1120},
  year={2005},
  organization={IEEE}
}

@article{geirhos2020shortcut,
  title={Shortcut learning in deep neural networks},
  author={Geirhos, Robert and Jacobsen, Jorn-Henrik and Michaelis, Claudio and Zemel, Richard and Brendel, Wieland and Bethge, Matthias and Wichmann, Felix A},
  journal={Nature Machine Intelligence},
  volume={2},
  number={11},
  pages={665--673},
  year={2020},
  publisher={Nature Publishing Group}
}

@inproceedings{selbst2018meaningful,
  title={“Meaningful Information” and the Right to Explanation},
  author={Selbst, Andrew and Powles, Julia},
  booktitle={Conference on Fairness, Accountability and Transparency},
  pages={48--48},
  year={2018},
  organization={PMLR}
}

@article{knight2019ai,
  title={AI and Machine Learning-Based Credit Underwriting and Adverse Action under the ECOA},
  author={Knight, Eric},
  journal={Bus. \& Fin. L. Rev.},
  volume={3},
  pages={236},
  year={2019},
  publisher={HeinOnline}
}

@article{covert2020understanding,
  title={Understanding global feature contributions with additive importance measures},
  author={Covert, Ian and Lundberg, Scott M and Lee, Su-In},
  journal={Advances in Neural Information Processing Systems},
  volume={33},
  pages={17212--17223},
  year={2020}
}

@article{shapley1953value,
  title={A value for n-person games},
  author={Shapley, Lloyd},
  journal={Contributions to the Theory of Games},
  number={28},
  pages={307--317},
  year={1953},
  publisher={Princeton University Press}
}

@inproceedings{binder2016layer,
  title={Layer-wise relevance propagation for neural networks with local renormalization layers},
  author={Binder, Alexander and Montavon, Gregoire and Lapuschkin, Sebastian and Muller, Klaus-Robert and Samek, Wojciech},
  booktitle={International Conference on Artificial Neural Networks},
  pages={63--71},
  year={2016},
  organization={Springer}
}

@inproceedings{datta2016algorithmic,
  title={Algorithmic transparency via quantitative input influence: Theory and experiments with learning systems},
  author={Datta, Anupam and Sen, Shayak and Zick, Yair},
  booktitle={2016 IEEE Symposium on Security and Privacy (SP)},
  pages={598--617},
  year={2016},
  organization={IEEE}
}

@inproceedings{sundararajan2017axiomatic,
  title={Axiomatic attribution for deep networks},
  author={Sundararajan, Mukund and Taly, Ankur and Yan, Qiqi},
  booktitle={International Conference on Machine Learning},
  pages={3319--3328},
  year={2017},
  organization={PMLR}
}

@inproceedings{kumar2020problems,
  title={Problems with Shapley-value-based explanations as feature importance measures},
  author={Kumar, I Elizabeth and Venkatasubramanian, Suresh and Scheidegger, Carlos and Friedler, Sorelle},
  booktitle={International Conference on Machine Learning},
  pages={5491--5500},
  year={2020},
  organization={PMLR}
}

@inproceedings{sundararajan2020many,
  title={The many Shapley values for model explanation},
  author={Sundararajan, Mukund and Najmi, Amir},
  booktitle={International Conference on Machine Learning},
  pages={9269--9278},
  year={2020},
  organization={PMLR}
}

@article{heskes2020causal,
  title={Causal Shapley values: Exploiting causal knowledge to explain individual predictions of complex models},
  author={Heskes, Tom and Sijben, Evi and Bucur, Ioan Gabriel and Claassen, Tom},
  journal={arXiv preprint arXiv:2011.01625},
  year={2020}
}

@article{chen2021explaining,
  title={Explaining a series of models by propagating local feature attributions},
  author={Chen, Hugh and Lundberg, Scott M and Lee, S},
  journal={CoRR abs/2105.00108},
  year={2021}
}

@inproceedings{wang2021shapley,
  title={Shapley flow: A graph-based approach to interpreting model predictions},
  author={Wang, Jiaxuan and Wiens, Jenna and Lundberg, Scott},
  booktitle={International Conference on Artificial Intelligence and Statistics},
  pages={721--729},
  year={2021},
  organization={PMLR}
}

@inproceedings{wang2020shapley,
  title={Shapley Explanation Networks},
  author={Wang, Rui and Wang, Xiaoqian and Inouye, David I},
  booktitle={International Conference on Learning Representations},
  year={2020}
}

@article{frye2020asymmetric,
  title={Asymmetric Shapley values: incorporating causal knowledge into model-agnostic explainability},
  author={Frye, Christopher and Rowat, Colin and Feige, Ilya},
  journal={Advances in Neural Information Processing Systems},
  volume={33},
  year={2020}
}

@article{vstrumbelj2014explaining,
  title={Explaining prediction models and individual predictions with feature contributions},
  author={{\v{S}}trumbelj, Erik and Kononenko, Igor},
  journal={Knowledge and Information Systems},
  volume={41},
  number={3},
  pages={647--665},
  year={2014},
  publisher={Springer}
}

@article{mitchell2021sampling,
  title={Sampling Permutations for Shapley Value Estimation},
  author={Mitchell, Rory and Cooper, Joshua and Frank, Eibe and Holmes, Geoffrey},
  journal={arXiv preprint arXiv:2104.12199},
  year={2021}
}

@article{covert2021explaining,
  title={Explaining by removing: A unified framework for model explanation},
  author={Covert, Ian and Lundberg, Scott and Lee, Su-In},
  journal={Journal of Machine Learning Research},
  volume={22},
  number={209},
  pages={1--90},
  year={2021}
}

@article{castro2017improving,
  title={Improving polynomial estimation of the Shapley value by stratified random sampling with optimum allocation},
  author={Castro, Javier and G{\'o}mez, Daniel and Molina, Elisenda and Tejada, Juan},
  journal={Computers \& Operations Research},
  volume={82},
  pages={180--188},
  year={2017},
  publisher={Elsevier}
}

@article{castro2009polynomial,
  title={Polynomial calculation of the Shapley value based on sampling},
  author={Castro, Javier and G{\'o}mez, Daniel and Tejada, Juan},
  journal={Computers \& Operations Research},
  volume={36},
  number={5},
  pages={1726--1730},
  year={2009},
  publisher={Elsevier}
}

@article{owen1972multilinear,
  title={Multilinear extensions of games},
  author={Owen, Guillermo},
  journal={Management Science},
  volume={18},
  number={5-part-2},
  pages={64--79},
  year={1972},
  publisher={INFORMS}
}

@inproceedings{okhrati2021multilinear,
  title={A multilinear sampling algorithm to estimate Shapley values},
  author={Okhrati, Ramin and Lipani, Aldo},
  booktitle={2020 25th International Conference on Pattern Recognition (ICPR)},
  pages={7992--7999},
  year={2021},
  organization={IEEE}
}

@article{dubey1981value,
  title={Value theory without efficiency},
  author={Dubey, Pradeep and Neyman, Abraham and Weber, Robert James},
  journal={Mathematics of Operations Research},
  volume={6},
  number={1},
  pages={122--128},
  year={1981},
  publisher={INFORMS}
}

@book{rubinstein2016simulation,
  title={Simulation and the Monte Carlo method},
  author={Rubinstein, Reuven Y and Kroese, Dirk P},
  volume={10},
  year={2016},
  publisher={John Wiley \& Sons}
}

@article{ruiz1998family,
  title={The family of least square values for transferable utility games},
  author={Ruiz, Luis M and Valenciano, Federico and Zarzuelo, Jose M},
  journal={Games and Economic Behavior},
  volume={24},
  number={1-2},
  pages={109--130},
  year={1998},
  publisher={Elsevier}
}

@article{deng1994complexity,
  title={On the complexity of cooperative solution concepts},
  author={Deng, Xiaotie and Papadimitriou, Christos H},
  journal={Mathematics of Operations Research},
  volume={19},
  number={2},
  pages={257--266},
  year={1994},
  publisher={INFORMS}
}

@article{faigle1992shapley,
  title={The Shapley value for cooperative games under precedence constraints},
  author={Faigle, Ulrich and Kern, Walter},
  journal={International Journal of Game Theory},
  volume={21},
  number={3},
  pages={249--266},
  year={1992},
  publisher={Springer}
}

@article{fatima2008linear,
  title={A linear approximation method for the Shapley value},
  author={Fatima, Shaheen S and Wooldridge, Michael and Jennings, Nicholas R},
  journal={Artificial Intelligence},
  volume={172},
  number={14},
  pages={1673--1699},
  year={2008},
  publisher={Elsevier}
}

@article{illes2019estimation,
  title={Estimation of the Shapley value by ergodic sampling},
  author={Ill{\'e}s, Ferenc and Ker{\'e}nyi, P{\'e}ter},
  journal={arXiv preprint arXiv:1906.05224},
  year={2019}
}

@article{granot2002cost,
  title={Cost allocation for a tree network with heterogeneous customers},
  author={Granot, Daniel and Kuipers, Jeroen and Chopra, Sunil},
  journal={Mathematics of Operations Research},
  volume={27},
  number={4},
  pages={647--661},
  year={2002},
  publisher={INFORMS}
}

@article{megiddo1978computational,
  title={Computational complexity of the game theory approach to cost allocation for a tree},
  author={Megiddo, Nimrod},
  journal={Mathematics of Operations Research},
  volume={3},
  number={3},
  pages={189--196},
  year={1978},
  publisher={INFORMS}
}

@inproceedings{jethani2021fastshap,
  title={FastSHAP: Real-Time {S}hapley Value Estimation},
  author={Jethani, Neil and Sudarshan, Mukund and Covert, Ian Connick and Lee, Su-In and Ranganath, Rajesh},
  booktitle={International Conference on Learning Representations},
  year={2021}
}

@inproceedings{simon2020projected,
  title={A Projected Stochastic Gradient Algorithm for Estimating Shapley Value Applied in Attribute Importance},
  author={Simon, Grah and Vincent, Thouvenot},
  booktitle={International Cross-Domain Conference for Machine Learning and Knowledge Extraction},
  pages={97--115},
  year={2020},
  organization={Springer}
}

@article{chen2018shapley,
  title={L-Shapley and C-Shapley: Efficient model interpretation for structured data},
  author={Chen, Jianbo and Song, Le and Wainwright, Martin J and Jordan, Michael I},
  journal={arXiv preprint arXiv:1808.02610},
  year={2018}
}

@phdthesis{maleki2015addressing,
  title={Addressing the computational issues of the Shapley value with applications in the smart grid},
  author={Maleki, Sasan},
  year={2015},
  school={University of Southampton}
}

@article{van2018new,
  title={A new approximation method for the Shapley value applied to the WTC 9/11 terrorist attack},
  author={van Campen, Tjeerd and Hamers, Herbert and Husslage, Bart and Lindelauf, Roy},
  journal={Social Network Analysis and Mining},
  volume={8},
  number={1},
  pages={1--12},
  year={2018},
  publisher={Springer}
}

@inproceedings{ghorbani2019data,
  title={Data Shapley: Equitable valuation of data for machine learning},
  author={Ghorbani, Amirata and Zou, James},
  booktitle={International Conference on Machine Learning},
  pages={2242--2251},
  year={2019},
  organization={PMLR}
}

@article{ghorbani2020neuron,
  title={Neuron Shapley: Discovering the responsible neurons},
  author={Ghorbani, Amirata and Zou, James},
  journal={arXiv preprint arXiv:2002.09815},
  year={2020}
}

@article{tarashev2016risk,
  title={Risk attribution using the Shapley value: Methodology and policy applications},
  author={Tarashev, Nikola and Tsatsaronis, Kostas and Borio, Claudio},
  journal={Review of Finance},
  volume={20},
  number={3},
  pages={1189--1213},
  year={2016},
  publisher={Oxford University Press}
}

@article{lucchetti2010shapley,
  title={The Shapley and Banzhaf values in microarray games},
  author={Lucchetti, Roberto and Moretti, Stefano and Patrone, Fioravante and Radrizzani, Paola},
  journal={Computers \& Operations Research},
  volume={37},
  number={8},
  pages={1406--1412},
  year={2010},
  publisher={Elsevier}
}

@article{moretti2010statistical,
  title={Statistical analysis of the Shapley value for microarray games},
  author={Moretti, Stefano},
  journal={Computers \& Operations Research},
  volume={37},
  number={8},
  pages={1413--1418},
  year={2010},
  publisher={Elsevier}
}

@article{tarashev2009systemic,
  title={The systemic importance of financial institutions},
  author={Tarashev, Nikola A and Borio, Claudio EV and Tsatsaronis, Kostas},
  journal={BIS Quarterly Review, September},
  year={2009}
}

@article{landinez2017shapley,
  title={Shapley Value: its algorithms and application to supply chains},
  author={Landinez-Lamadrid, Daniela C and Ramirez-Rios, Diana G and Neira Rodado, Dionicio and Parra Negrete, Kevin Armando and Combita Nino, Johana Patricia},
  journal={INGE CUC},
  year={2017},
  publisher={Corporaci{\'o}n Universidad de la Costa}
}

@incollection{aumann1994economic,
  title={Economic applications of the Shapley value},
  author={Aumann, Robert JJ},
  booktitle={Game-theoretic methods in general equilibrium analysis},
  pages={121--133},
  year={1994},
  publisher={Springer}
}

@inproceedings{kapishnikov2019xrai,
  title={Xrai: Better attributions through regions},
  author={Kapishnikov, Andrei and Bolukbasi, Tolga and Vi{\'e}gas, Fernanda and Terry, Michael},
  booktitle={Proceedings of the IEEE/CVF International Conference on Computer Vision},
  pages={4948--4957},
  year={2019}
}

@article{ren2021learning,
  title={Learning Baseline Values for Shapley Values},
  author={Ren, Jie and Zhou, Zhanpeng and Chen, Qirui and Zhang, Quanshi},
  journal={arXiv preprint arXiv:2105.10719},
  year={2021}
}

@inproceedings{williamson2020efficient,
  title={Efficient nonparametric statistical inference on population feature importance using Shapley values},
  author={Williamson, Brian and Feng, Jean},
  booktitle={International Conference on Machine Learning},
  pages={10282--10291},
  year={2020},
  organization={PMLR}
}

@article{vstrumbelj2009explaining,
  title={Explaining instance classifications with interactions of subsets of feature values},
  author={{\v{S}}trumbelj, Erik and Kononenko, Igor and {\v{S}}ikonja, M Robnik},
  journal={Data \& Knowledge Engineering},
  volume={68},
  number={10},
  pages={886--904},
  year={2009},
  publisher={Elsevier}
}

@article{lipovetsky2001analysis,
  title={Analysis of regression in game theory approach},
  author={Lipovetsky, Stan and Conklin, Michael},
  journal={Applied Stochastic Models in Business and Industry},
  volume={17},
  number={4},
  pages={319--330},
  year={2001},
  publisher={Wiley Online Library}
}

@incollection{charnes1988extremal,
title={Extremal principle solutions of games in characteristic function form: core, {C}hebychev and {S}hapley value generalizations},
author={Charnes, A and Golany, B and Keane, M and Rousseau, J},
booktitle={Econometrics of Planning and Efficiency},
pages={123--133},
year={1988},
publisher={Springer}
}

@article{ethayarajh2021attention,
  title={Attention flows are Shapley value explanations},
  author={Ethayarajh, Kawin and Jurafsky, Dan},
  journal={arXiv preprint arXiv:2105.14652},
  year={2021}
}

@article{rozemberczki2022shapley,
  title={The Shapley Value in Machine Learning},
  author={Rozemberczki, Benedek and Watson, Lauren and Bayer, P{\'e}ter and Yang, Hao-Tsung and Kiss, Oliv{\'e}r and Nilsson, Sebastian and Sarkar, Rik},
  journal={arXiv preprint arXiv:2202.05594},
  year={2022}
}

@article{karczmarz2021improved,
  title={Improved Feature Importance Computations for Tree Models: Shapley vs. Banzhaf},
  author={Karczmarz, Adam and Mukherjee, Anish and Sankowski, Piotr and Wygocki, Piotr},
  journal={arXiv preprint arXiv:2108.04126},
  year={2021}
}

@inproceedings{chen2020ls,
  title={Ls-tree: Model interpretation when the data are linguistic},
  author={Chen, Jianbo and Jordan, Michael},
  booktitle={Proceedings of the AAAI Conference on Artificial Intelligence},
  volume={34},
  number={04},
  pages={3454--3461},
  year={2020}
}

@inproceedings{koh2020concept,
  title={Concept bottleneck models},
  author={Koh, Pang Wei and Nguyen, Thao and Tang, Yew Siang and Mussmann, Stephen and Pierson, Emma and Kim, Been and Liang, Percy},
  booktitle={International Conference on Machine Learning},
  pages={5338--5348},
  year={2020},
  organization={PMLR}
}

@article{monderer2002variations,
  title={Variations on the {S}hapley value},
  author={Monderer, Dov and Samet, Dov and others},
  journal={Handbook of Game Theory},
  volume={3},
  pages={2055--2076},
  year={2002},
  publisher={North-Holland Mannheim}
}

@book{van2000asymptotic,
  title={Asymptotic statistics},
  author={Van der Vaart, Aad W},
  volume={3},
  year={2000},
  publisher={Cambridge University Press}
}

@article{covert2022learning,
  title={Learning to Estimate Shapley Values with Vision Transformers},
  author={Covert, Ian and Kim, Chanwoo and Lee, Su-In},
  journal={arXiv preprint arXiv:2206.05282},
  year={2022}
}

@inproceedings{reiter2020developing,
  title={Developing an interpretable schizophrenia deep learning classifier on fMRI and sMRI using a patient-centered DeepSHAP},
  author={Reiter, Jacob},
  booktitle={in 32nd Conference on Neural Information Processing Systems (NeurIPS 2018)(Montreal: NeurIPS)},
  pages={1--11},
  year={2020}
}

@article{miroshnikov2021mutual,
  title={Mutual information-based group explainers with coalition structure for machine learning model explanations},
  author={Miroshnikov, Alexey and Kotsiopoulos, Konstandinos and Kannan, Arjun Ravi},
  journal={arXiv preprint arXiv:2102.10878},
  year={2021}
}

@book{aumann2015values,
  title={Values of non-atomic games},
  author={Aumann, Robert J and Shapley, Lloyd S},
  year={2015},
  publisher={Princeton University Press}
}

\appendix

\section{Appendix}

\subsection{Datasets}

In order to compare the unbiased stochastic estimators and their variants, we utilize three datasets with varying numbers of features.

\subsubsection{Diabetes}

The diabetes dataset ($n=442$) consists of ten input
% baseline
features (e.g., age, sex, BMI, etc.) and a continuous output which is diabetes disease progression measured one year after measuring baseline features.  

\subsubsection{NHANES}
\label{sec:data:nhanes}

The NHANES (National Health and Nutrition Examination Survey) (I) dataset ($n=14264$) consists of 79 input
% baseline
features (e.g., age, sex, BMI, etc.) and a binary output for which the positive label is 5-year mortality after measuring the patient features.
% baseline features.

\subsubsection{Blog}

The blog dataset ($n=52397$) consists of 280 features (e.g., number of comments, length of blog post, etc.) and a non-binary output which is the number of comments in the next twenty-four hours.  

\subsection{Experiments}

% \hugh{TODO: Incorporate pseduocode for each model agnostic approach?}

\begin{figure}[!ht]
\includegraphics[width=\textwidth]{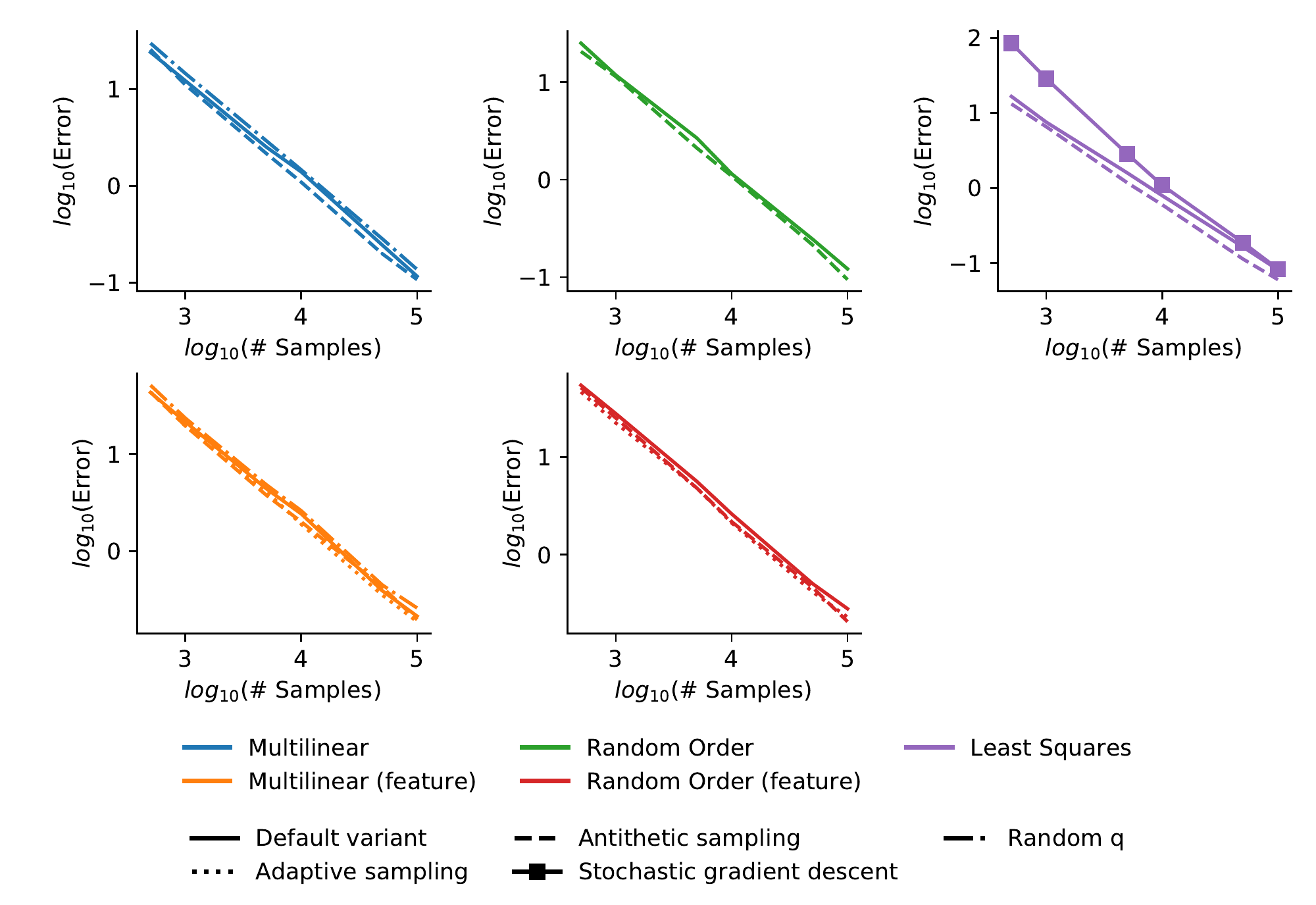}
\centering
\caption{Errors for all variants of unbiased stochastic estimators in the diabetes dataset.}
\label{fig:diabetes_all_error}
\end{figure}

\begin{figure}[!ht]
\includegraphics[width=\textwidth]{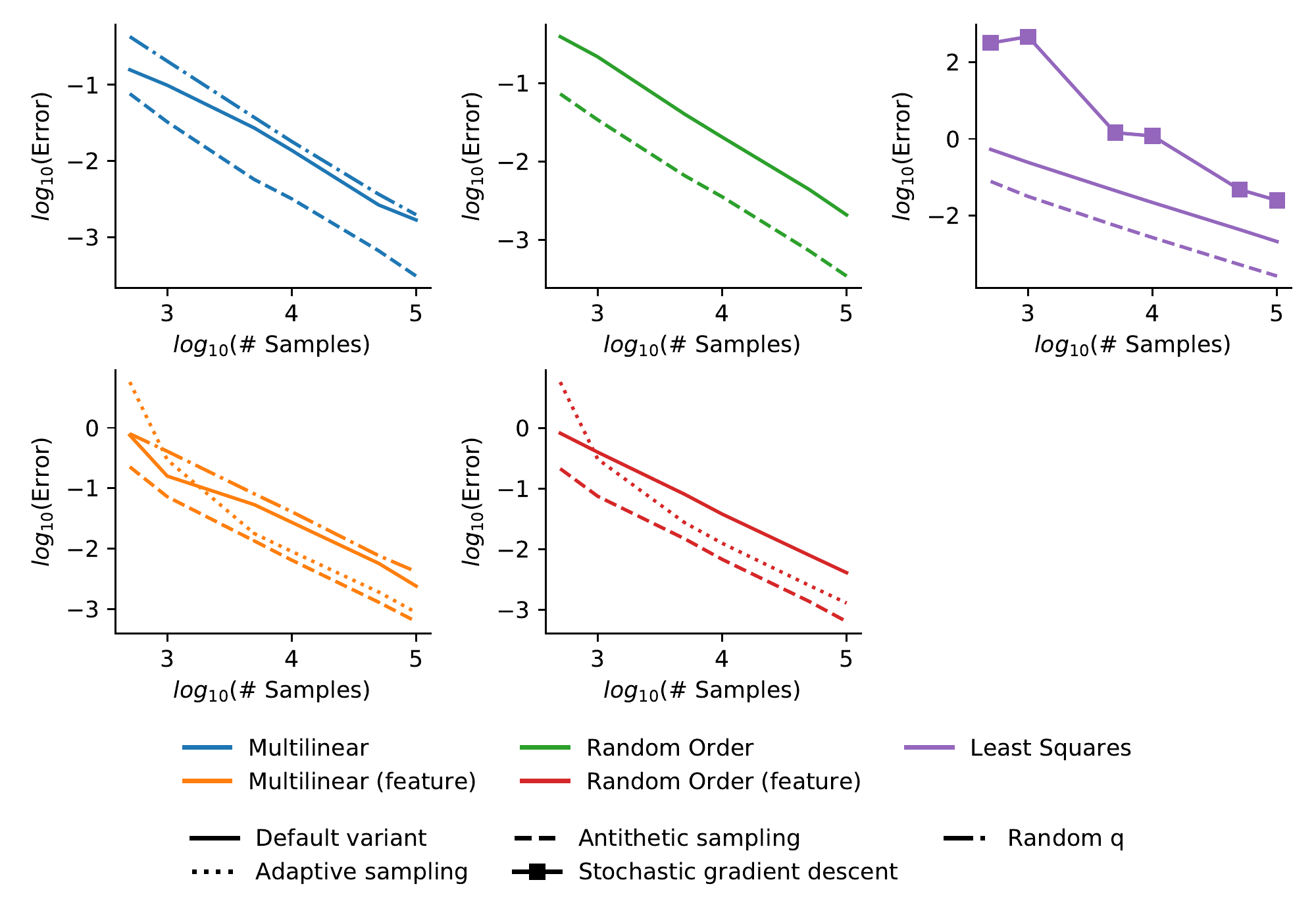}
\centering
\caption{Errors for all variants of unbiased stochastic estimators in the NHANES dataset.}
\label{fig:nhanes_all_error}
\end{figure}

\begin{figure}[!ht]
\includegraphics[width=\textwidth]{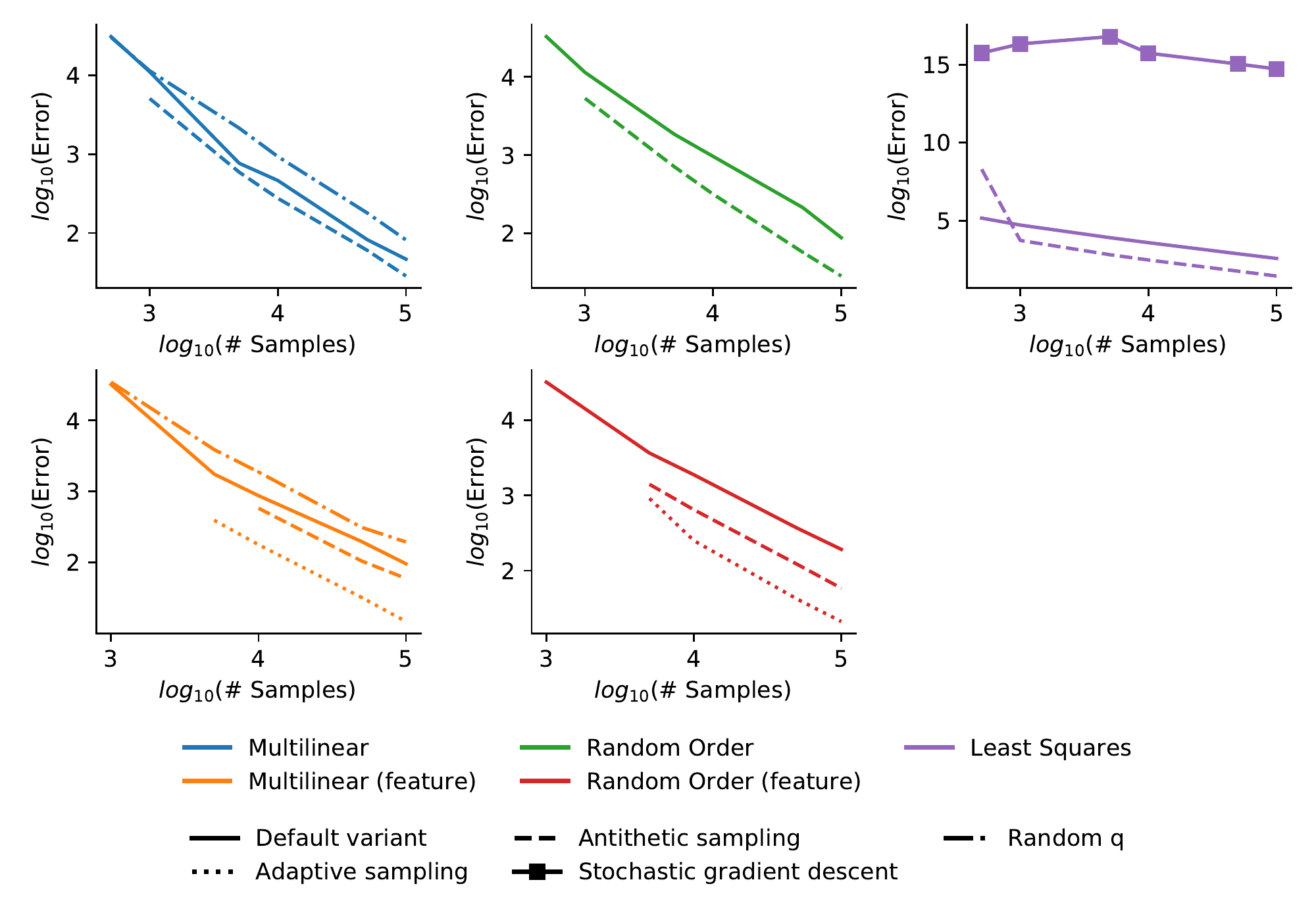}
\centering
\caption{Errors for all variants of unbiased stochastic estimators in the blog dataset.}
\label{fig:blog_all_error}
\end{figure}

\begin{figure}[!ht]
\includegraphics[width=.7\textwidth]{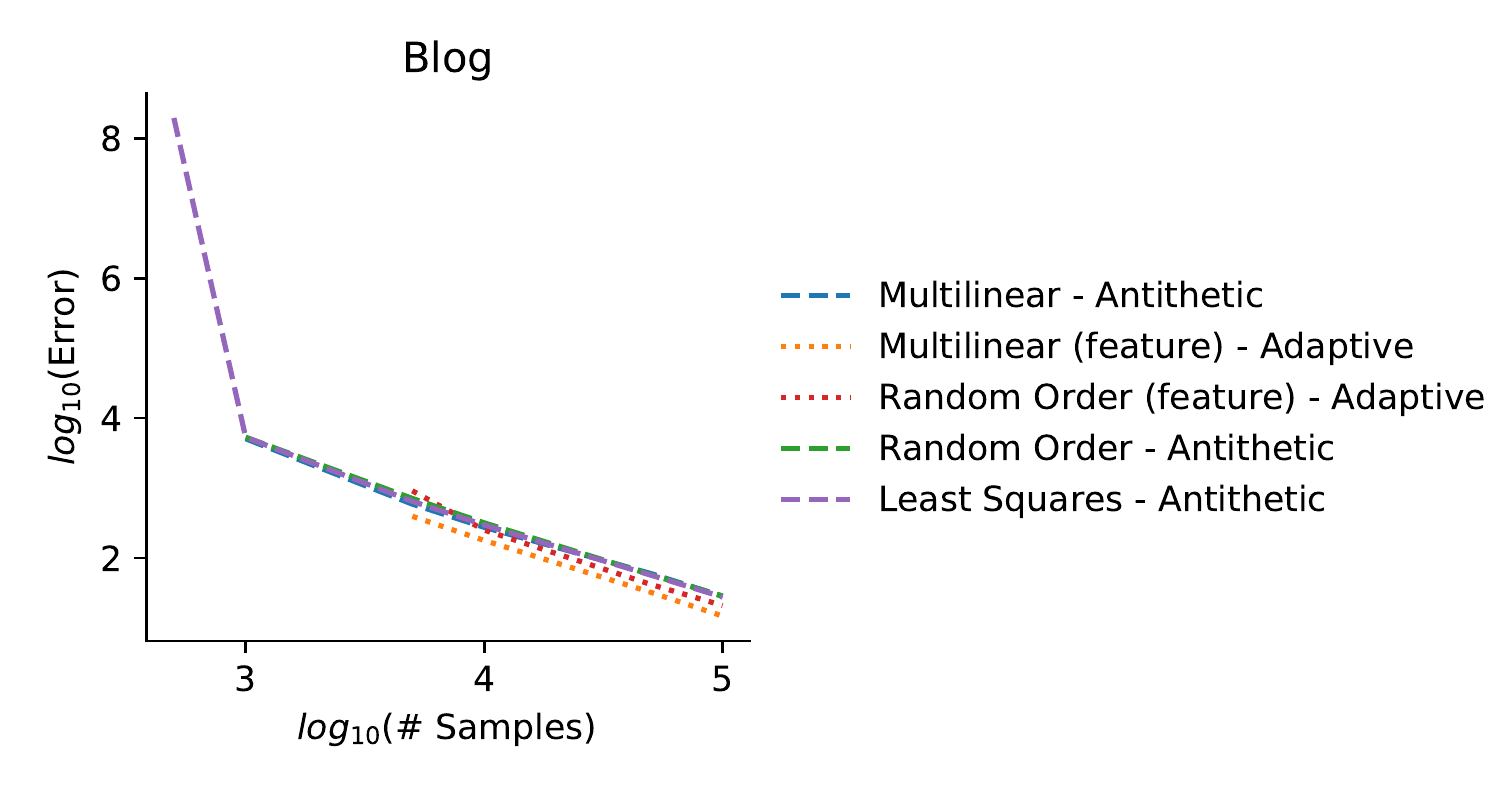}
\centering
\caption{Untruncated comparison of best variants on the blog dataset.}
\label{fig:blog_top_error}
\end{figure}

\begin{figure}[!ht]
\includegraphics[width=\textwidth]{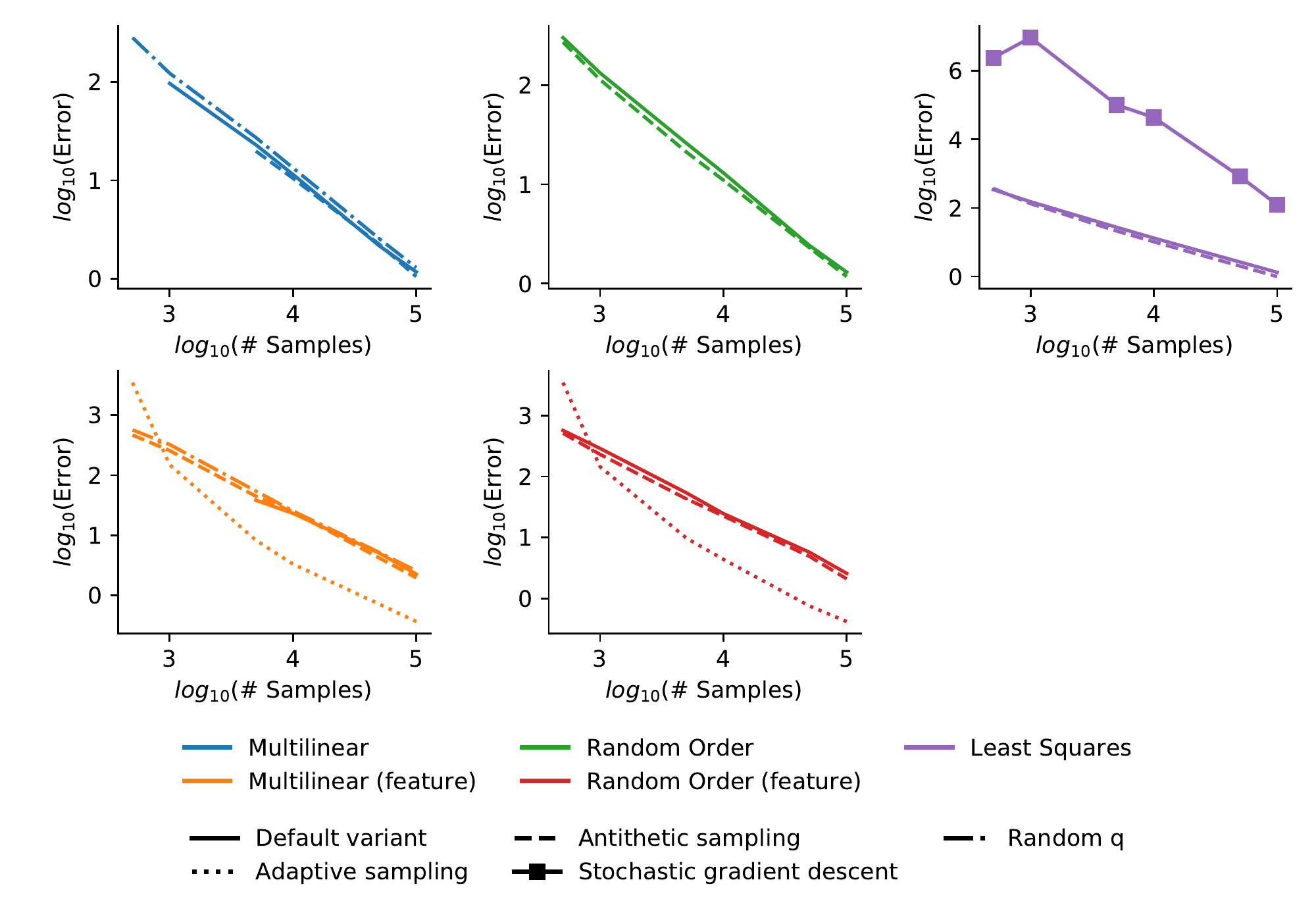}
\centering
\caption{Errors for all variants of unbiased stochastic estimators in the diabetes dataset (with 100 additional zero features).}
\label{fig:diabetes_zeros_all_error}
\end{figure}

\begin{figure}[!ht]
\includegraphics[width=.7\textwidth]{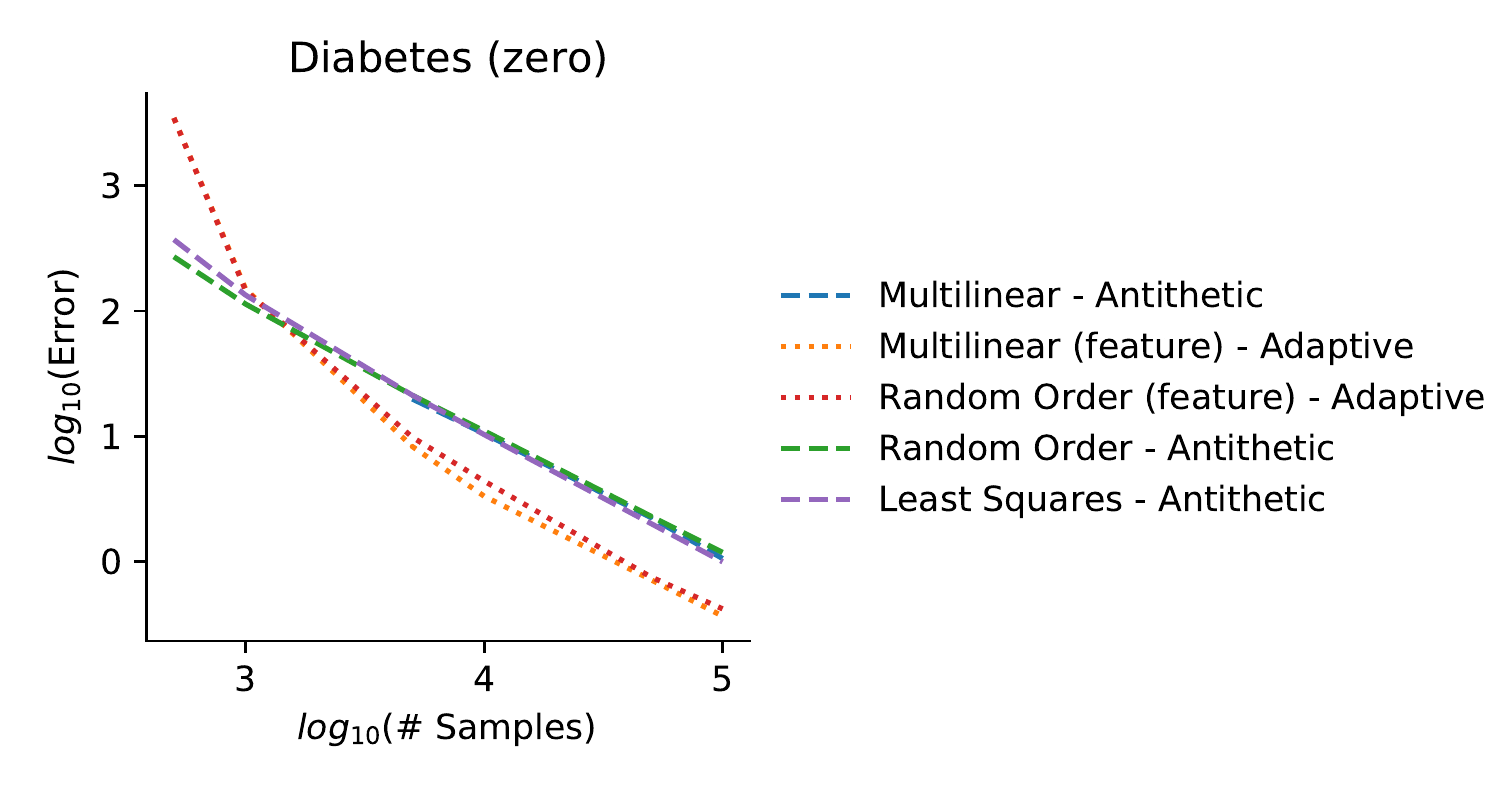}
\centering
\caption{Comparison of best variants in the diabetes dataset with 100 additional zero features.}
\label{fig:diabetes_zeros_top_error}
\end{figure}

In Figures \ref{fig:diabetes_all_error}-\ref{fig:diabetes_zeros_top_error}, we show more comprehensive comparisons of the various estimators' error based on mean squared error between estimated baseline Shapley values and the true baseline Shapley values estimated by TreeSHAP.  See Figure~\ref{fig:shapley_algorithms} for a description of each method.  We include two new variants: ``Random q'' is multilinear extension sampling with uniformly drawn $q$ and ``Stochastic Gradient Descent'' is SGD-Shapley~\cite{simon2020projected}, which is only applicable to the least squares estimator.  Note that missing values denote that we had insufficient numbers of samples (coalitions) to create an estimate of the feature attribution for all features.  

In Tables \ref{tab:diabetes_bias}-\ref{tab:blog_var}, we show the bias and variance of each approach, which sum to the estimator error~\cite{covert2020improving}.  We refer to the estimators using shorthand in the tables.
In particular, the stochastic estimators include each sampling-based approach, where the prefixes MEF (multilinear extension, feature-wise), ROF (random order, feature-wise), RO (random order), and LS (least squares) correspond to multilinear extension sampling~\cite{okhrati2021multilinear}, IME~\cite{strumbelj2010efficient}, ApproShapley~\cite{castro2009polynomial}, and KernelSHAP~\cite{lundberg2017unified} respectively.  The suffixes ``ANTI'' and ``ADAPT'' stand for antithetic sampling and adaptive sampling respectively.  Finally, ``SGD'' is SGD-Shapley and ``RAND'' is multilinear extension sampling with uniformly drawn $q$.

These results largely mirror the results in Figure \ref{fig:shapley_algorithms}.  Antithetic and adaptive sampling are largely helpful.  In the diabetes dataset, which has a small number of features, antithetic and adaptive sampling are only mildly helpful.  For NHANES, which has a medium number of features, there is a larger separation between approaches and antithetic sampling is more helpful than adaptive sampling.  For the blog dataset, which has a large number of features, adaptive sampling is more beneficial than antithetic sampling.  For the multilinear approaches, we find that using a random $q$ naturally produces results very similar to the random order sampling approaches.  This is natural because they are actually equivalent approaches to sampling subsets from the appropriate distribution $P(S)$.  However, we do see that the default version of multilinear sampling benefits over random $q$ by sampling $q$ at fixed intervals according to the trapezoid rule.  Finally, we see that for the least squares estimators, using SGD is unhelpful.  Note that some approaches have non-zero bias although they are provably unbiased.  This is because of the number of trials performed.  If we performed more trials, their bias would continue to shrink.

\begin{table}[!ht]
    \centering
    \begin{tabular}{|l|l|l|l|l|l|l|}
    \hline
        ~ & 500 & 1000 & 5000 & 10000 & 50000 & 100000 \\ \hline
        ME & 1.03307 & 0.10672 & 0.00882 & 0.01493 & 0.0019 & 0.00103 \\ \hline
        ME\_RAND & 0.14194 & 0.16995 & 0.03135 & 0.01725 & 0.00251 & 0.00062 \\ \hline
        ME\_ANTI & 2.20398 & 0.61293 & 0.0586 & 0.03112 & 0.00139 & 0.00072 \\ \hline
        MEF & 1.1593 & 0.93785 & 0.04937 & 0.01506 & 0.00507 & 0.00246 \\ \hline
        MEF\_RAND & 0.31587 & 0.21718 & 0.01556 & 0.0228 & 0.0015 & 0.00604 \\ \hline
        MEF\_ADAPT & 8.92679 & 1.18728 & 0.13272 & 0.03672 & 0.0013 & 0.00155 \\ \hline
        MEF\_ANTI & 10.47164 & 2.51886 & 0.13555 & 0.03785 & 0.00179 & 0.00254 \\ \hline
        RO & 0.25328 & 0.13665 & 0.04269 & 0.00863 & 0.00248 & 0.00268 \\ \hline
        RO\_ANTI & 0.07361 & 0.16267 & 0.01586 & 0.01166 & 0.00082 & 0.00059 \\ \hline
        ROF & 0.15102 & 0.22011 & 0.03267 & 0.047 & 0.00572 & 0.00221 \\ \hline
        ROF\_ADAPT & 0.46465 & 0.22019 & 0.05365 & 0.01299 & 0.00403 & 0.00348 \\ \hline
        ROF\_ANTI & 0.47356 & 0.21328 & 0.01301 & 0.01196 & 0.00784 & 0.00207 \\ \hline
        LS & 0.29439 & 0.06221 & 0.00499 & 0.00546 & 0.00323 & 0.00047 \\ \hline
        LS\_ANTI & 0.18394 & 0.15229 & 0.00925 & 0.00485 & 0.00255 & 0.00062 \\ \hline
        LS\_SGD & 43.17269 & 13.02032 & 0.50849 & 0.116 & 0.00464 & 0.00135 \\ \hline
    \end{tabular}
\caption{Biases in the diabetes dataset.}
\label{tab:diabetes_bias}
\end{table}

\begin{table}[!ht]
    \centering
    \begin{tabular}{|l|l|l|l|l|l|l|}
    \hline
        ~ & 500 & 1000 & 5000 & 10000 & 50000 & 100000 \\ \hline
        ME & 0.01634 & 0.00219 & 0.0002 & 0.00026 & 3e-05 & 4e-05 \\ \hline
        ME\_RAND & 0.00337 & 0.00371 & 0.00031 & 0.00025 & 7e-05 & 2e-05 \\ \hline
        ME\_ANTI & 0.00061 & 0.01537 & 0.00019 & 8e-05 & 1e-05 & 0.0 \\ \hline
        MEF & 0.00686 & 0.01644 & 0.00051 & 0.0003 & 3e-05 & 2e-05 \\ \hline
        MEF\_RAND & 0.00444 & 0.00152 & 0.00035 & 0.00037 & 7e-05 & 3e-05 \\ \hline
        MEF\_ADAPT & ~ & 0.00611 & 0.00076 & 0.0002 & 1e-05 & 1e-05 \\ \hline
        MEF\_ANTI & 0.00248 & 0.00041 & 0.00146 & 0.00029 & 2e-05 & 0.0 \\ \hline
        RO & 0.00096 & 0.00196 & 0.00014 & 0.00013 & 2e-05 & 1e-05 \\ \hline
        RO\_ANTI & 0.00041 & 0.00016 & 0.0001 & 1e-05 & 0.0 & 0.0 \\ \hline
        ROF & 0.00396 & 0.0044 & 0.00054 & 0.0004 & 8e-05 & 2e-05 \\ \hline
        ROF\_ADAPT & ~ & 0.00516 & 0.00013 & 0.00015 & 3e-05 & 1e-05 \\ \hline
        ROF\_ANTI & 0.003 & 0.00047 & 0.00012 & 7e-05 & 1e-05 & 0.0 \\ \hline
        LS & 0.00441 & 0.00193 & 0.00032 & 0.00022 & 3e-05 & 2e-05 \\ \hline
        LS\_ANTI & 0.00188 & 0.00039 & 5e-05 & 2e-05 & 0.0 & 0.0 \\ \hline
        LS\_SGD & 2.40455 & 4.0216 & 0.00907 & 0.00703 & 0.00045 & 0.00019 \\ \hline
    \end{tabular}
\caption{Biases in the NHANES dataset.}
\label{tab:nhanes_bias}
\end{table}

\begin{table}[!ht]
    \centering
    \begin{tabular}{|l|l|l|l|l|l|l|}
    \hline
        ~ & 500 & 1000 & 5000 & 10000 & 50000 & 100000 \\ \hline
        ME & 338.01236 & 178.15781 & 144.77503 & 13.6816 & 1.86531 & 0.59309 \\ \hline
        ME\_RAND & 610.90174 & 99.46345 & 24.6845 & 17.25537 & 6.09554 & 1.5173 \\ \hline
        ME\_ANTI & ~ & 83.27304 & 573.95994 & 89.70199 & 1.58388 & 0.80663 \\ \hline
        MEF & ~ & 395.61402 & 592.00777 & 69.757 & 11.14579 & 4.95535 \\ \hline
        MEF\_RAND & ~ & 486.29512 & 113.96514 & 35.53342 & 2.63154 & 7.78507 \\ \hline
        MEF\_ADAPT & ~ & ~ & 719.24301 & 108.48492 & 5.07824 & 0.71839 \\ \hline
        MEF\_ANTI & ~ & ~ & ~ & 555.99106 & 10.3927 & 3.22064 \\ \hline
        RO & 811.14593 & 360.14181 & 142.15739 & 45.46918 & 5.03847 & 2.11807 \\ \hline
        RO\_ANTI & ~ & 268.88612 & 20.02851 & 4.23618 & 2.26818 & 0.74835 \\ \hline
        ROF & ~ & 941.03632 & 66.14242 & 24.57762 & 13.38569 & 2.99248 \\ \hline
        ROF\_ADAPT & ~ & ~ & 79.202 & 23.71196 & 1.32897 & 0.94973 \\ \hline
        ROF\_ANTI & ~ & ~ & 29.90393 & 13.26942 & 1.75937 & 0.58583 \\ \hline
        LS & 5140.67576 & 1640.50381 & 207.70321 & 116.66386 & 20.67612 & 10.15152 \\ \hline
        LS\_ANTI & 5589054.21794 & 525.46966 & 29.46567 & 10.70906 & 1.70923 & 0.78873 \\ \hline
        LS\_SGD & 1.4363e+14 & 6.1849e+14 & 1.8168e+15 & 1.5086e+14 & 2.9198e+13 & 1.4884e+13 \\ \hline
    \end{tabular}
\caption{Biases in the blog dataset.}
\label{tab:blog_bias}
\end{table}

\begin{table}[!ht]
    \centering
    \begin{tabular}{|l|l|l|l|l|l|l|}
    \hline
        ~ & 500 & 1000 & 5000 & 10000 & 50000 & 100000 \\ \hline
        ME & 22.96666 & 12.07279 & 2.5081 & 1.37924 & 0.2432 & 0.11547 \\ \hline
        ME\_RAND & 29.62088 & 14.31613 & 2.89015 & 1.44624 & 0.28177 & 0.13646 \\ \hline
        ME\_ANTI & 23.43875 & 10.42177 & 2.09982 & 1.06178 & 0.19789 & 0.10684 \\ \hline
        MEF & 41.77143 & 20.47749 & 4.35186 & 2.41331 & 0.38869 & 0.21297 \\ \hline
        MEF\_RAND & 50.69169 & 23.15461 & 4.82135 & 2.58893 & 0.44981 & 0.25554 \\ \hline
        MEF\_ADAPT & 34.51499 & 18.23867 & 3.99338 & 1.86179 & 0.35407 & 0.18755 \\ \hline
        MEF\_ANTI & 32.8124 & 16.97314 & 3.63817 & 1.9772 & 0.42555 & 0.19555 \\ \hline
        RO & 24.84541 & 11.66307 & 2.63066 & 1.13767 & 0.24184 & 0.1187 \\ \hline
        RO\_ANTI & 20.58076 & 11.24632 & 2.09441 & 1.07522 & 0.21244 & 0.09355 \\ \hline
        ROF & 53.78942 & 27.42756 & 5.58406 & 2.5637 & 0.5001 & 0.27848 \\ \hline
        ROF\_ADAPT & 45.63946 & 21.95396 & 4.75253 & 2.10057 & 0.41285 & 0.22687 \\ \hline
        ROF\_ANTI & 50.41685 & 24.72881 & 4.7859 & 2.1644 & 0.45592 & 0.205 \\ \hline
        LS & 16.28934 & 7.47362 & 1.5866 & 0.78921 & 0.16094 & 0.08113 \\ \hline
        LS\_ANTI & 12.98929 & 6.39006 & 1.16766 & 0.59549 & 0.11076 & 0.0598 \\ \hline
        LS\_SGD & 41.51592 & 15.59966 & 2.33451 & 0.97674 & 0.1831 & 0.08241 \\ \hline
    \end{tabular}
\caption{Variances in the diabetes dataset.}
\label{tab:diabetes_var}
\end{table}

\begin{table}[!ht]
    \centering
    \begin{tabular}{|l|l|l|l|l|l|l|}
    \hline
        ~ & 500 & 1000 & 5000 & 10000 & 50000 & 100000 \\ \hline
        ME & 0.10806 & 0.07575 & 0.02124 & 0.01065 & 0.00206 & 0.00129 \\ \hline
        ME\_RAND & 0.33544 & 0.15789 & 0.02918 & 0.01401 & 0.00282 & 0.00154 \\ \hline
        ME\_ANTI & 0.05972 & 0.01036 & 0.00433 & 0.00244 & 0.00051 & 0.00024 \\ \hline
        MEF & 0.59095 & 0.10908 & 0.04178 & 0.02123 & 0.00447 & 0.00187 \\ \hline
        MEF\_RAND & 0.62686 & 0.32213 & 0.06354 & 0.03194 & 0.00598 & 0.00325 \\ \hline
        MEF\_ADAPT & ~ & 0.22775 & 0.01328 & 0.0069 & 0.00149 & 0.00068 \\ \hline
        MEF\_ANTI & 0.17626 & 0.05679 & 0.00915 & 0.00481 & 0.00101 & 0.00049 \\ \hline
        RO & 0.30973 & 0.16971 & 0.03165 & 0.01606 & 0.00345 & 0.00163 \\ \hline
        RO\_ANTI & 0.05721 & 0.02674 & 0.00511 & 0.00278 & 0.00057 & 0.00027 \\ \hline
        ROF & 0.64611 & 0.31183 & 0.06334 & 0.02974 & 0.00625 & 0.00323 \\ \hline
        ROF\_ADAPT & ~ & 0.23807 & 0.0216 & 0.00981 & 0.00198 & 0.00102 \\ \hline
        ROF\_ANTI & 0.16582 & 0.05885 & 0.01168 & 0.00532 & 0.00109 & 0.00051 \\ \hline
        LS & 0.41585 & 0.19099 & 0.03514 & 0.01713 & 0.00339 & 0.00165 \\ \hline
        LS\_ANTI & 0.05979 & 0.02459 & 0.0043 & 0.00209 & 0.00042 & 0.00021 \\ \hline
        LS\_SGD & 247.45077 & 361.76017 & 1.1384 & 0.9365 & 0.03746 & 0.01957 \\ \hline
    \end{tabular}
\caption{Variances in the NHANES dataset.}
\label{tab:nhanes_var}
\end{table}

\begin{table}[!ht]
    \centering
    \begin{tabular}{|l|l|l|l|l|l|l|}
    \hline
        ~ & 500 & 1000 & 5000 & 10000 & 50000 & 100000 \\ \hline 
        ME & 86410.36336 & 30866.41722 & 1997.75462 & 1281.98781 & 227.58978 & 129.44555 \\ \hline
        ME\_RAND & 84274.19531 & 31874.46777 & 5954.03044 & 2590.77554 & 493.77726 & 226.55744 \\ \hline
        ME\_ANTI & ~ & 14121.88171 & 1076.49978 & 680.50752 & 166.27577 & 78.60651 \\ \hline
        MEF & ~ & 89335.0449 & 4330.56682 & 2375.37671 & 542.53634 & 264.14241 \\ \hline
        MEF\_RAND & ~ & 96454.13874 & 10778.75215 & 5232.66527 & 873.32245 & 537.40648 \\ \hline
        MEF\_ADAPT & ~ & ~ & 385.15507 & 391.50072 & 85.38466 & 40.68452 \\ \hline
        MEF\_ANTI & ~ & ~ & ~ & 1071.56372 & 284.39153 & 164.1287 \\ \hline
        RO & 90722.92558 & 31639.72416 & 5028.47504 & 2659.61346 & 596.35393 & 245.73287 \\ \hline
        RO\_ANTI & ~ & 14649.27372 & 1965.69685 & 886.23589 & 159.37583 & 79.43453 \\ \hline
        ROF & ~ & 89654.02281 & 10234.68309 & 5268.76653 & 1020.95547 & 534.06821 \\ \hline
        ROF\_ADAPT & ~ & ~ & 2479.70421 & 679.7336 & 115.98881 & 57.75736 \\ \hline
        ROF\_ANTI & ~ & ~ & 3914.76444 & 1797.94149 & 339.38264 & 162.71561 \\ \hline
        LS & 3.9848e5 & 1.4281e5 & 22220.79433 & 10600.85713 & 2062.52593 & 1032.15956 \\ \hline
        LS\_ANTI & 5.4101e+8 & 14374.92093 & 1777.1319 & 819.92753 & 157.75298 & 76.68178 \\ \hline
        LS\_SGD & 1.5715e+16 & 5.9717e+16 & 1.7423e+17 & 1.5233e+16 & 3.0639e+15 & 1.4746e+14 \\ \hline
    \end{tabular}
\caption{Variances in the blog dataset.}
\label{tab:blog_var}
\end{table}

\end{document}